\documentclass[acmsmall]{acmart}
\AtBeginDocument{%
  \providecommand\BibTeX{{%
    \normalfont B\kern-0.5em{\scshape i\kern-0.25em b}\kern-0.8em\TeX}}}

\usepackage{amsmath, amsfonts}
\usepackage[ruled]{algorithm2e}
\usepackage{algorithmic}
\usepackage{multirow}
\usepackage{textcomp}
\usepackage{xcolor}
\usepackage{tikz}
\usetikzlibrary{matrix, calc}
\usepackage{color, colortbl}

\setcopyright{acmcopyright}       
\copyrightyear{XX}               
\acmYear{XX}                     
\acmDOI{10.1145/3712712}

\acmJournal{THRI}                  
\acmVolume{XX}                      
\acmNumber{XX}                       
\acmArticle{XX}                    
\acmMonth{0}                        

\begin{document}

\title{From Novice to Skilled: RL-based Shared Autonomy Communicating with Pilots in UAV Multi-Task Missions}

\author{Kal Backman}
\email{Kal.Backman@monash.edu}
\orcid{0000-0002-6898-4945}                      
\author{Dana Kulić}
\email{Dana.Kulic@monash.edu}               
\orcid{0000-0002-4169-2141}                     
\author{Hoam Chung}
\email{Hoam.Chung@monash.edu}
\orcid{0000-0001-7044-3729}                      
\affiliation{%
  \institution{Monash University}
  \country{Australia}
}

\renewcommand{\shortauthors}{K. Backman et al.}
\newcommand{\norm}[1]{\left\lVert#1\right\rVert}

\begin{abstract}
Multi-task missions for unmanned aerial vehicles (UAVs) involving inspection and landing tasks are challenging for novice pilots due to the difficulties associated with depth perception and the control interface. We propose a shared autonomy system, alongside supplementary information displays, to assist pilots to successfully complete multi-task missions without any pilot training. Our approach comprises of three modules: (1) a perception module that encodes visual information onto a latent representation, (2) a policy module that augments pilot's actions, and (3) an information augmentation module that provides additional information to the pilot. 
The policy module is trained in simulation with simulated users and transferred to the real world without modification in a user study ($\mathbf{n=29}$), alongside alternative supplementary information schemes including learnt red/green light feedback cues and an augmented reality display. The pilot’s intent is unknown to the policy module and is inferred from the pilot’s input and UAV’s states.  
The assistant increased task success rate for the landing and inspection tasks from [16.67\% \& 54.29\%] respectively to [95.59\% \& 96.22\%]. With the assistant, inexperienced pilots achieved similar performance to experienced pilots. Red/green light feedback cues reduced the required time by 19.53\% and trajectory length by 17.86\% for the inspection task, where participants rated it as their preferred condition due to the intuitive interface and providing reassurance. 
This work demonstrates that simple user models can train shared autonomy systems in simulation, and transfer to physical tasks to estimate user intent and provide effective assistance and information to the pilot.
\end{abstract}




\maketitle

\section{Introduction}
Unmanned aerial vehicles (UAVs) are renowned for their mobility, often deployed in search and rescue \cite{ThermalHumanRescue, MountainRescue, SARDO} and inspection \cite{BoilerInspection, BridgeInspection, PVInspection} tasks due to their ability to manoeuvre in full 3D space. However this manoeuvrability comes at a cost of increased teleoperation complexity due to difficulties associated with pilots’ depth perception  \cite{StereoscopicFPV, Kal2} and control dexterity. Due to these challenges it is difficult for novice pilots to successfully complete the aforementioned tasks. 

Autonomous solutions have been proposed for such tasks \cite{BoilerInspection, BridgeInspection, PVInspection}, however often require that the structure of the environment be known a priori, or contain a set of fixed, known mission objectives. The main limitation of fully autonomous solutions is their inability to replicate high-level human decision making and general artificial intelligence \cite{Robots4HarshEnv, HumanVsRobotInspection}. Therefore teleoperation control schemes are preferred in real-life UAV operations over fully autonomous solutions \cite{SemiAutoSnR} to take advantage of high-level human decision making.

However, teleoperation of such complex missions typically requires expert pilots. Prior works demonstrate that operator error is a major factor in UAV accidents \cite{joslin2015synthesis, fernando2017survey}.
If assistance strategies can be introduced to minimize the requisite piloting expertise required to safely complete complex tasks, drone technologies could be used more widely by domain experts (e.g., ecologists and structural engineers) without the need for extensive training or hiring dedicated expert pilots. 

Assistance strategies include the use of shared autonomy, which combines the control inputs of human pilots with that of artificial intelligence to collaboratively complete a set of objectives. Three main challenges arise when developing shared autonomy systems: inferring the intent of the user, providing control outputs to complete the inferred objective and deciding how and what information should be communicated back to the user. 

The first challenge, inferring the intent of the user can be performed implicitly by observing the user’s actions within the context of the observable environment, or explicitly by requesting the user to specify their objective. Although inferring intent implicitly poses the risk of incorrect goal estimation leading to a misalignment of objectives between the AI and user, users prefer implicit intent estimation methods due to their intuitiveness and reduction in cognitive workload \cite{Kal2, ImplicitVisualIntent}. 

For the second challenge, the automated assistant must deliver its control outputs considering its uncertainty about the user’s intent. Acting too early risks taking an incorrect action not aligned with the user, while waiting to build sufficient confidence in the user’s intent before acting can lead to delayed assistance and task failure.
Further issues arise with how much control should the assistant exert over the system. Providing insufficient assistance can lead to task failure while excessive control deteriorates team effectiveness in collaborative tasks \cite{MutualAdaptation}. 

For the third challenge, communication feedback promotes transparency of the shared autonomy system, providing increased observability and predictability of system behaviour \cite{TransparencyInSA}. However developing natural feedback communication channels that do not hinder a user’s control input capabilities, prevent loss of focus from context switching and are designed for environments with high auditory noise is difficult.  

Of the limited shared autonomy works, none yet deal with multi-task missions, which are missions that require the sequential completion of objectives, and few consider single task missions where the goal (e.g., landing location) must be inferred. Prior UAV shared autonomy works instead focus on single goal inspection \cite{SharedAutonomyPoleInspection} or landing \cite{Reddy} tasks, or are restricted to obstacle avoidance for use in inspection tasks \cite{SemiAutoSnR, HapticUAVInspection}. For non-UAV shared autonomy works, multiple ambiguous goal inference has been demonstrated, however prior works require that all potential goals are defined a priori and are assumed to be visible and static at start of operation \cite{jeon2020shared, javdani2018shared, yow2023shared}. These restrictions limit the practicality of such approaches in large unstructured environments due to the difficulty in numerically defining goals and the inability to attain such priors as they are not known by the user before operation, such as in the context of structural defect and wildlife detection missions.

Our prior works \cite{Kal1, Kal2} proposed a shared autonomy solution capable of providing assistance under ambiguity of the pilot’s intent in unknown environments, but were limited to landing tasks and a restricted control scheme where the UAV’s yaw is locked. Our prior experiments indicated that participants were able to significantly improve task performance, however sometimes struggled to discern the intent of the assistant or what was expected of them as a pilot, where lack of communication feedback led to a sense of uncertainty. Therefore in this paper we propose a system capable of UAV shared autonomy for multi-task missions, while introducing learnt feedback communication and information augmentation to further enhance user experience and task performance.

The contributions of this work are the following: (1) we introduce multi-task missions to shared autonomy UAV systems, providing assistance under uncertainty of both goal and objective; (2) we propose a methodology for learning to communicate with human pilots, that is learnt by interacting with simulated users; (3) we demonstrate an approach to increase exploration throughput by mitigating the need for high visual fidelity simulation environments whilst maintaining the robustness to real-life deployment by jointly mapping high- and low-quality images onto identical latent embeddings; (4) we also contribute further improvements to the TD3 \cite{TD3} reinforcement learning algorithm by sharing and jointly optimising network components which is shown to be essential for both task success and communication with the pilot.

The proposed work is capable of delivering beneficial assistance to pilots where the pilot’s objective, intent and environment are not known a priori, allowing the system to be transferred to new operational environments. The system is trained purely on simulated data, removing the need for human supervision during training which is time costly and hazardous for the robot and human overseer. Despite being trained on synthetic data with simulated users, the model is directly transferrable to the physical domain without modification, able to assist both novice and expert human pilots to achieve equal performance. The approach simultaneously provides task-relevant assistance and communication feedback by jointly learning the two components into a single model, assisting naïve pilots who are unaware of the inner workings of the assistant and without pre-establishing operation conventions or user preferences.

\section{Related work}
\subsection{Autonomous Drone Inspection} A number of prior works propose algorithms for autonomous drone inspection in specific industrial \cite{PlaneInspection} and infrastructure \cite{PVInspection, InspectionViewPoint, EMInspection, lyu2023vision} inspection tasks. Methods used by autonomous inspection works to generate trajectories include leveraging the structure of the environment such as parallel rows of solar panels \cite{PVInspection} or the planar structure of building façades \cite{lyu2023vision}. Works such as \cite{InspectionViewPoint} focused on maximizing image quality by proposing an optimal viewpoint recommendation system to account for target size and sun illumination. 
However such works are limited to being either validated in simulation, dependent on the specific structure of the environment or limited to a small immediately visible search area. 

\subsection{Shared Autonomy Drone Inspection} 
Previous shared autonomy works for UAV inspection related tasks are limited.  \citet{SharedAutonomyPoleInspection} demonstrate a physical UAV for pole-like object inspections by transforming pilot inputs into cylindrical coordinates. 
\citet{SemiAutoSnR} implement a semi-autonomous UAV for GPS denied search and rescue missions using the pilot’s inputs to create waypoints for an autonomous safe trajectory path planner to follow. 
\citet{HapticUAVInspection} flew a simulated UAV through a virtual forest-like environment using haptic feedback and the closest safe action to the pilot’s input computed from a control barrier function.
The previous shared autonomy approaches are limited in handling ambiguity in the pilot’s intent, where \citet{SharedAutonomyPoleInspection} focused on single goal environments, while \citet{SemiAutoSnR, HapticUAVInspection} are delegated to obstacle avoidance tasks irrespective of the pilot’s objective. The prior works have yet to be validated in physical user studies, where \citet{SharedAutonomyPoleInspection, SemiAutoSnR} performed physical demonstrations while \citet{HapticUAVInspection} hosted a simulated user study.

\subsection{Autonomous Drone Landing}
Similar to inspection related tasks, many prior works related to UAV landing focus on fully autonomous solutions over shared autonomy. Deep-reinforcement learning methods for UAV landing have been proposed in \cite{SDQN, EmergencyLanding, MovingPlatformDDPG, MovingPlatformDDPG2}, where \citet{SDQN} used a sequence of deep Q-networks to search and land on physical targets. 
\citet{EmergencyLanding} used PPO to autonomously detect and land at undefined safe landing zones for physical emergency landing situations.
Both works \citet{SDQN, EmergencyLanding} are limited to a set of discrete action outputs, limiting the manoeuvrability of the UAV.
 
Continuous action approaches have been demonstrated in \cite{MovingPlatformDDPG, MovingPlatformDDPG2}, where \citet{MovingPlatformDDPG} used DDPG \cite{DDPG} to land a physical drone on a moving platform by outputting a target velocity along the horizontal plane at a constant descent.
Similarly, \citet{MovingPlatformDDPG2} used DDPG to land on a simulated moving platform by outputting a target velocity along the horizontal plane and a simple heuristic model to control for vertical descent.
Both \citet{MovingPlatformDDPG, MovingPlatformDDPG2} require the position of the target platform to be directly observable within their respective input state spaces and are limited to horizontal movement to counter the instability of continuous action space reinforcement learning.  

\subsection{Shared Autonomy Drone Landing}
Shared autonomy UAV landing approaches alleviate the need to explicitly define the landing zone due to the AI assistant inferring the pilot’s goal from observations of the their actions. \citet{Reddy} implemented a shared autonomy system to assist users in safely landing in the lunar lander game where the location of safe landing zones is only known to the user. The system uses a model free DQN to provide assistance to participants in a user study ($n=12$). However when applied to a physical drone ($n=4$), the landing pad location is included within the network’s state space and is limited to a discrete set of locations. 

Our initial work \cite{Kal1} demonstrated a shared autonomy system to assist pilots landing a simulated UAV on one of several platforms. The assistant comprised of two modules; a perception module responsible for encoding visual data onto a compressed latent representation and a policy module which was trained on simulated users using the reinforcement algorithm TD3 \cite{TD3} to augment the pilot’s actions. The assistant was not aware of the pilot’s true goal nor the structure of the environment, where the pilot’s goal must be inferred from observations of their actions in the context of the observed environment. The approach was validated on human pilots ($n=33$) in a simulated environment and demonstrated in a physical environment. 

Our subsequent work \cite{Kal2} further built on \cite{Kal1} and was evaluated in a physical user study. The assistant in \cite{Kal2} comprised of a perception and policy module which were both trained purely on synthetic data. 
Compared to \cite{Kal1}, the perception module fused two downwards facing RGB-D cameras onto a consolidated latent representation using a novel camera projection model that is only attainable in simulation for greater field of view. 
The policy module introduced an LSTM-cell to address the inconsistency concerns participants had in \cite{Kal1}, where policy convergence time and performance were significantly improved by providing hidden information of the simulated user’s intent to the critic only. 
The simulated user’s parametric model was expanded to take into account pilots' joystick thumb movements and task complexity was increased due to additional safe landing criteria and a greater diversity in environment structures.
The assistant successfully transferred to reality and was validated in a physical user study with human pilots ($n=28$) where it increased landing success rate from 51.4\% in the unassisted condition to 98.2\% in the assisted condition. Participants’ stated that their most preferred aspect was the intuitive design, however the non-intrusive behaviour made it difficult to discern the intent of the assistant or what was expected of them as a pilot, where the lack of communication feedback led to a sense of uncertainty.

As of yet there are no prior works to the authors’ knowledge that enable shared autonomy control of UAVs for multi-task missions. 

\subsection{Robotic Communication and Supplementary Information}
Prior works that communicate robot intent or augment the operator’s perception by providing supplementary information can be found in \cite{walker2018communicating, hedayati2018improving, brooks2020visualization, brock2018flymap, aleotti2017detection, szafir2015communicating, herdel2021drone, bevins2021aerial}. Common approaches use augmented reality ocular glasses such as the HoloLens to overlay information onto the physical world. 
\citet{walker2018communicating} demonstrates such an approach to highlight the intent and state of a UAV for a shared resource task where the UAV is given priority over the human participant. The UAV’s intent is conveyed through the display of navigation points, planned trajectory arrows or a virtual floating eye such that participants can observe its gaze. \citet{walker2018communicating} uses a mini map to allow participants to constantly perceive the physical state of the UAV relative to them. 
\citet{hedayati2018improving} similarly uses an augmented reality device for an inspection task requiring participants to take photos of targets as quickly as possible whilst maximizing the coverage of the target within the image. \citet{hedayati2018improving} augments the pilot’s perception by displaying the video feed through various methods. 
\citet{brooks2020visualization} aims to help convey the intent of a robotic manipulator to participants in a shared control robotic grasping task. The system’s goal belief is virtually rendered as a sphere and the future target end effector poses are visualized to highlight its intended trajectory.

Supplementary information can also be provided without ocular glasses. \citet{brock2018flymap} propose to display supplementary information without the use of ocular glasses by using an onboard projector to project a map of the robot's surroundings onto the ground. \citet{aleotti2017detection} augments a monitor display to show UAV state information and visualizes detected radioactivity levels using a 3D histogram to assist pilots in locating radioactive sources. The advantage of implementing augmented reality interfaces, either with or without glasses, is the ability to display a diverse range of high-quality visualizations.

An alternative to augmented reality interfaces is the use of onboard methods.  \citet{szafir2015communicating} use onboard LEDs to communicate the planned travel direction or the current heading of the UAV. 
\citet{herdel2021drone} introduces an anthropomorphized simulated UAV capable of displaying simplified human facial expressions through an onboard digital screen. \citet{bevins2021aerial} explores human perception of physical UAV movement patterns in communicating the intent of the UAV and how humans would respond to such movements.
The advantage of onboard communication methods is the simplicity in their design compared to augmented reality which requires complex localization and synchronization techniques to ensure the virtual projection aligns with reality.
Onboard techniques such as LEDs are inexpensive, easily retrofitted to existing drones as demonstrated in \citet{szafir2015communicating} whilst being a familiar interface for experienced pilots who rely on LEDs to perceive the state of the flight controller.

\subsection{Non-UAV Shared Autonomy}
Shared autonomy has also been explored for non-UAV robotic systems \cite{jeon2020shared, javdani2018shared, yow2023shared}. \citet{jeon2020shared} develop a shared autonomy assistant to aid participants to manipulate a robotic arm to prepare a culinary dish which requires multiple tasks to be performed and goals inferred. \citet{jeon2020shared} maps participant input onto a lower dimensional latent vector that is conditional on the current state and belief in the participant’s goal. The participant’s latent action is decoded and then blended with the assistant’s action across all of the goal beliefs similar to \citet{javdani2018shared}. Like \citet{javdani2018shared}, \citet{jeon2020shared} requires that all goals are explicitly known to the assistant as a priori. The assistant is unable to perceive the environment but instead is reliant on copying expert demonstrations in an identical environment in which they were demonstrated.

\citet{yow2023shared} demonstrates shared autonomy for robotic arm object gripping tasks in a cluttered scene leading to high ambiguity in inferring the goal of the participant. \citet{yow2023shared} further expands \citet{javdani2018shared} by modeling the problem as a discrete action POMDP, allowing the robot to deploy hindsight optimization actions \cite{javdani2018shared} whilst also being capable of performing single goal-oriented actions or not providing any assistance when appropriate. 
Further innovations are demonstrated by enabling the robot to use verbal cues posed as yes or no questions that explicitly confirm the participant’s intent to reduce ambiguity.
Similar to \citet{jeon2020shared} and \citet{javdani2018shared}, \citet{yow2023shared} requires that all potential goals be defined prior to operation. \citet{yow2023shared} improves this process by autonomously segmenting a point-cloud image captured at the beginning of operation. However this process requires human intervention for cluttered scenes and only works for static environments where all objects can be observed at the beginning from a single image.

\section{Problem Statement and Model Overview}
\begin{figure}[b]
\centering
\includegraphics[width=1.0\columnwidth]{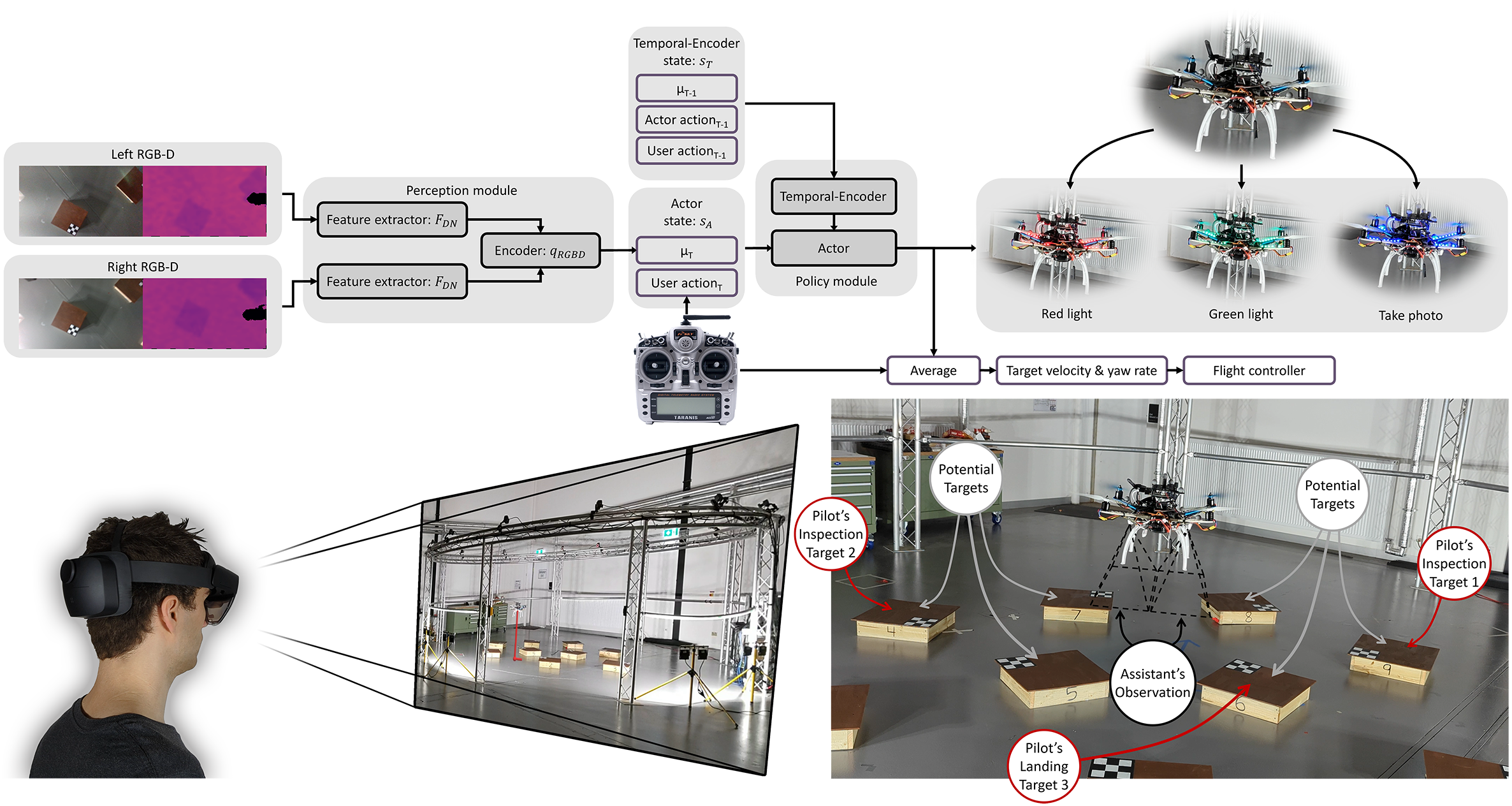}
\caption{System overview. (Top) Assistant network architecture. (Bottom left) Information augmentation using Microsoft HoloLens. (Bottom right) Example of a potential multi-task mission. }
\label{SystemOverview}
\end{figure}

We consider an unknown environment in which the UAV is operating. The environment contains several points of interest to the pilot (potential goals) where a task is to be performed. The onboard assistant is unaware of the global structure of the environment, capable of only sensing the local area with onboard sensors. The assistant is unaware of the pilot’s goal, which must be deciphered from  the pilot’s joystick inputs in the context of the observable environment. Together with a goal estimate, the assistant is required to estimate and provide assistance with the type of task the pilot needs to perform on the goal. To further enhance teamwork, the assistant can not only provide task specific controls that influence the physical state of the UAV, but can also communicate with the pilot by providing indication about the potential success of the task or to prompt them to take a desired action. 

Our approach is summarised in Fig.~\ref{SystemOverview} and is comprised of three components: (i) a perception module that is responsible for compressing high-dimensional visual information onto a latent representation, (ii) a policy module that is responsible for providing control and communicative outputs to assist the pilot in successfully completing tasks, and (iii) an information augmentation module that provides additional information to the pilot in discerning the state of the UAV. The perception module and the assistant are trained purely on synthetic data, learning to provide assistance to complete multi-task missions with simulated users based on a parametric model.

Multi-task missions comprise of landing and inspection tasks on points of interest named \emph{platforms}. These platforms are indistinguishable from one another and have a defined orientation, which is marked by a small checkerboard pattern at the top-left corner. To achieve success for an inspection task, an image must be captured where all four corners of the goal platform are visible to the downwards facing cameras, with the image correctly aligned to the platform below a certain yaw threshold. A landing task is considered successful given the UAV remains at rest atop of the goal platform while the UAV is correctly aligned to the platform below a certain yaw threshold.

\section{Perception module}\label{PerceptionModule}
The purpose of the perception module is to allow the assistant to visually perceive the environment so that potential goal targets can be identified. The perception module aims to do this by encoding high-dimensional visual information onto a low-dimensional latent vector for use in the policy module’s input state-space. As the policy module accumulates experience by interacting with the simulated user within the environment in real time, it is costly to explore large state-spaces, therefore a reduction in the dimensionality of the input state-space leads to accelerated policy convergence \cite{DimensionalityReduction}.

Learning from the low-dimensional mean latent vector \(\mu\) aims to reduce policy convergence times by reducing the time required to the explore the input state-space. However, given a data input \(x_k\) which is derived from sensor readings that capture a high-dimensional and complex system such as cameras, artificially generating \(x_k\) in simulation becomes computationally expensive. From the data efficiency point of view, it is preferred that data generated in simulation avoid computationally expensive high-frequency details and capture only the raw essence of what they are intended to represent. However, learning from simplistic simulated data sources is not indicative of the complexity and diversity of real life data sources and can later cause the learned policy to fail to bridge the sim-to-real gap. In order to maximise both the computational efficiency of learning from simple data source simulations and the robustness in transferring to reality with high-detailed and diversified data sources, we aim to encode both data sources onto an identical latent representation such that they can be used interchangeably.

Given a data source \(\bar{x}_k\) which represents the basic low-detail information of \(x_k\), encoder \(q_k\) maps both \(\bar{x}_k\) and \(x_k\) onto a normally distributed latent vector with means \(\bar{\mu}\) \& \(\mu\) and variances \(\bar{\sigma}^2\) \& \(\sigma^2\) respectively. As both \(\bar{x}_k\) and \(x_k\) represent the same observation but in varying detail, both latent samples \(\bar{z} \sim \mathcal{N}(\bar{\mu}, \bar{\sigma}^2)\) \& \(z \sim \mathcal{N}(\mu, \sigma^2)\) are reconstructed using decoder \(p_l\) to generate a simplified representation of \(y_l\) which is denoted as \(\bar{y}_l\). Both \(\bar{x}_k\) and \(x_k\) are reconstructed to \(\bar{y}_l\) and not \(\bar{y}_l\) \& \(y_l\) respectively due to \(\bar{x}_k\) not containing the high-detail information required to accurately reconstruct \(y_l\), which would result in both latent representations \(\bar{\mu}\) \& \(\mu\) not being identical. To further ensure consistency between latent representations \(\bar{\mu}\) \& \(\mu\), the cosine similarity loss is introduced as a regularizing term such that both latent vectors are co-linear with each other. The introduction of the cosine similarity loss transforms the CM-VAE architecture into a cross-modal similarity variational auto-encoder (CM-SVAE).

Information regarding the implementation of the CM-SVAE architecture can be found in Appendix \ref{CM-SVAE_Implementation}.

\section{Policy module}\label{PolicyModelSection}
The purpose of the policy module is to assist pilots in successfully completing multi-task missions from a set of two unique tasks, inspection and landing. The policy module achieves this by outputting a target velocity and yaw rate that is subsequently averaged with the pilot's current joystick velocity and yaw rate input. This joint signal is the command input to the drone flight controller. The policy module can provide additional information to pilots through the use of feedback cues by turning on red or green lights attached to the UAV. We formulate our problem as a partially observable Markov decision process (POMDP) where the set of all possible states is defined as \(\mathcal{S}\). The state transition processes \(T\) is treated as a stochastic process due to the difficulty in modeling UAV dynamics from unobservable forces resulting from the ground effect: \(T : \mathcal{S} \times \mathcal{A} \times \mathcal{S} \rightarrow [0, 1]\), where \(\mathcal{A}\) is the set of all actions that can be taken by the policy module. The policy module observes the environment using the observation function \(\mathcal{O}\), where the pilot's goal and intent are treated as hidden states which must be inferred from successive observations: \(\mathcal{O} : \mathcal{S} \times \Omega \rightarrow [0, 1]\) where \(\Omega\) denotes the set of observations. The reward function is defined as \(R : \mathcal{S} \times \mathcal{A} \times \mathcal{S} \rightarrow \mathbb{R}\), where the aim of reinforcement learning is to find the closest approximation to the optimal policy: \(\pi : \mathcal{S} \times \mathcal{A}\) which maximises the expected future reward with discount factor \(\gamma \in [0,1]\). 

For optimal policy approximation we build upon our prior works \cite{Kal1, Kal2} and implement TD3 \cite{TD3} but with additional novel contributions to promote shared learning between the actor and critic. TD3 is chosen as a base algorithm due to providing model-free continuous action space control whilst alleviating the instability concerns of its predecessor DDPG \cite{DDPG}. TD3 is an off-policy, policy gradient, actor-critic approach that uses the minimum value of the two critics in estimating the Q-value for the next state action pair, the full algorithm is described in Algorithm 1 in \cite{TD3}.

From our prior work \cite{Kal2}, it was found that the greatest influence of the policy network’s performance was providing the critic with the simulated user’s hidden state, the goal landing platform location. This additional information only attainable during training in simulation can supplement the critic’s state space, transforming the POMDP problem into a fully observable Markov decision process (MDP) for the critic only. As MDPs present a simpler problem to POMDPs, the critic was able to learn the task requirements quicker and at a greater proficiency. Using this principle we aim to further improve TD3 \cite{TD3}  by instilling knowledge from the critic to the actor through shared networks that both actor and critic can jointly optimise. We also introduce supervised learning to the reinforcement learning problem for these shared networks to promote desired actor behavioural characteristics. We call this novel approach Shared-TD3. 

\subsection{Simulated Environment} \label{TD3Implementation}
For modelling UAV dynamics we use the Unreal Engine plugin Airsim \cite{AirSim} as our simulation environment. Compared to our previous works \cite{Kal1} and \cite{Kal2}, the simulated UAVs fly in an empty Airsim environment devoid of textures, lighting and platforms, where all graphical rendering is offloaded to the OpenGL renderer described in Appendix \ref{CM-SVAE_Implementation} for increased computational efficiency. Platforms are generated in the OpenGL rendering environment by randomly placing \(N_p\) platforms with random orientations such that adjacent platforms are not within a minimum threshold distance while the platforms width, length and height are randomly sampled. At each time step the actor only has access to the latent representation encoded by the perception module as described in section~\ref{PerceptionModule}, the simulated user’s input action and the actor’s previous output action.   

\subsection{Simulating Users}\label{PolicyModelSimUser}
To train the policy module we use simulated users to generate potential actions human pilots could take.  Previously \cite{Kal2} we defined simulated users based on a parametric model containing four parameters: \(\alpha\), \(\beta\), \(\Psi\) \& \( \Phi\). \(\alpha \in [0,1]\) describes the simulated user’s conformance to the assistant’s actions and models how likely a pilot will adopt the policy of the assistant. \(\beta \in [0,1]\) describes the simulated user’s depth perception proficiency, the ability to improve one’s estimate of the current goal platform’s position and orientation. Both \(\alpha\) and \(\beta\) influence the simulated user’s current estimate of the goal position and orientation by: 
\begin{equation}
\hat{G}_{i+1} = \hat{G}_i + \alpha \frac{a_a - a_u}{K_\alpha} + \beta \frac{G - \hat{G}_i}{K_\beta}, \label{GoalUpdateEq}
\end{equation}
where \(a_a\) \& \(a_u\) are the previous action taken by the assistant and simulated user respectively, while \(K_\alpha\) \& \(K_\beta\) are scaling constants. 

\(\Psi\) and \(\Phi\) aim to model pilots' thumb movements on a joystick controller where \(\Psi \in [0,1]\) describes how aggressively a pilot accelerates their thumbs on the joystick controller while \(\Phi \in [0,1]\) determines how far on the joystick they will push, controlling for maximum desired flight speed. The direction in which the simulated user flies is determined by setting waypoints using a state machine comprising of three states: approach, descent and inspection state. The simulated user initially aligns the yaw to a set orientation, after which it will set waypoints to travel to based on the current estimate of the goal location \(\hat{G}_i\) and which state it is currently in.

To account for the introduction of feedback cues from the assistant turning on red and green lights, simulated users contain a new state variable “patience” \(\in [0,1]\) and are able to enter a “red” or “green” state in response to the assistant’s feedback. The default state of the simulated user is the “no-light” state, to enter the red or green state \(N\) consecutive timesteps where only the respective light has been displayed is required. The simulated user will remain in the red or green state given the respective light continues to be displayed. The simulated user will return to the default no-light state given \(M\) timesteps without the respective light being displayed or the simulated user’s patience has expired. The simulated user’s patience decreases given a timestep where the red or green light is displayed in proportion to the simulated user’s \(\alpha\) parameter and increases when no light is displayed in proportion to \(\alpha\). Once a simulated user’s patience decreases to zero they are considered frustrated and remain in the frustrated state until their patience has returned to the maximum value of one. 

Whilst in the red state simulated users will exert less control over the UAV by having their output action \(a_u\) scaled by \(S_r \in [0.0, 0.5]\) to become smaller in magnitude, allowing the assistant to more easily take control of the UAV.
Whilst in the green state the simulated user will initiate actions that are essential to completing the objective. If the current task is an inspection task, entering the green state will cause the simulated user to immediately request a picture to be taken. If the current task is to land, then the simulated user will start to descend. By causing the simulated user to enter the green state the assistant is able to quickly progress the current task and receive the task completion reward in fewer state transitions. 
To represent a broader range of potential human pilots, samples of simulated users are assigned to either: ignore all feedback and will not enter the red / green state, have a hard requirement to be in the green state to progress the mission or not having a hard requirement for the green state but still able to enter the red / green state.

When training the assistant, new populations of simulated users are sampled after each epoch by independently generating each of the four parameters \(\alpha\), \(\beta\), \(\Psi\) \& \( \Phi\) from a uniform distribution. Two seeds for random number generators which control deterministic and non-deterministic characteristics described in \cite{Kal2} are sampled, including the simulated user's aforementioned response to light feedback.
Examples of simulated users flying the UAV in various generated environments can be seen in the supplementary video. 

\subsection{Shared-TD3}
In our prior work \cite{Kal2}, the actor-critic networks followed a multi-branch configuration where the first network branch focused on extracting information from the current state \(s_t\) whilst the second branch extracted temporal information using the prior states \(s_{t-n}\). The two branch’s outputs were concatenated and fed through the output layers to produce an action or Q-value for the respective actor-critic networks. Our shared TD3 approach follows our previous approach using multi-branch actor-critic networks.  However, each actor and critic shares an identical network weight temporal branch which is referred to as the Temporal-Encoder. There are two main advantages of introducing the Temporal-Encoder: reduction in computation time per training iteration by reusing the Temporal-Encoder’s LSTM state for calculating both the actor and critic’s output and a reduction in total training iterations required for the actor to learn a suitable policy due to the critic propagating additional gradients through the network.  

To further condition the LSTM cell’s temporal embedding, an additional network is introduced, the Temporal-Decoder which is trained in a supervised manner. The objective of the Temporal-Decoder is to take the output of the Temporal-Encoder and estimate key features that can only be attained from subsequent state observations across time such as the time since the pilot has requested an inspection image. The advantage of introducing supervised learning is the ability to forcefully embed information required for task completion onto the LSTM embedding without the need for the actor to explore and learn the representation by itself, which is not guaranteed using reinforcement learning. 

\subsubsection{Shared-TD3 Algorithm}\label{Shared-TD3 Algorithm}
The training procedure for shared TD3 initially samples a mini-batch of \(N\) state transitions \( (s, a, r, s^\prime) \) and an additional ground truth state \(y\) for the Temporal-Decoder to estimate. State \(s\) comprises of three components \(s=[s_T, s_A, s_C]\) which corresponds to the Temporal-Encoder input of \(n\) time lags (\(s_T\)), actor current branch input state  (\(s_A\)) and the critic current branch input state  (\(s_C\)). Network optimisation is split into three steps: 1) Temporal-Decoder, 2) critic then 3) actor optimisation. For Step~1, the temporal state \(s_T\) is passed into the Temporal-Encoder \(q_T\), which is subsequently fed into the Temporal-Decoder \(p_T\) to produce an estimate of the current output values i.e. \(\hat{y} = p_T(q_T(s_T))\). The error between the estimate \(\hat{y}\) and \(y\) is computed using loss function \(f_L\) where each of the respective elements in \(y\) are summed then are averaged across mini-batch of size \(N\): \(Loss = N^{-1} \Sigma (\Sigma f_L(y_i, \hat{y}_i)) \). Gradients are then propagated through \(p_T\) \& \(q_T\) with their respective weights updated. 

For Step~2, critic optimisation, the next temporal LSTM state embedding \(x^\prime_{LSTM}\) is calculated using the target temporal encoder: \(x^\prime_{LSTM} = q^\prime_T(s^\prime_T) \). The next temporal LSTM state embedding is subsequently fed into the target actor network \(\pi^\prime\) alongside the actor’s next current state input \(s^\prime_A\) from which clipped gaussian noise is added to produce the target actor’s action for the next state: \(a^\prime = \pi^\prime(x^\prime_{LSTM}, s^\prime_A) + \epsilon\), where \(\epsilon \sim \)clip\((\mathcal{N}(\mu, \sigma^2), -c, c)\). \(a^\prime\) is then concatenated with the critic’s next current state input \(s^\prime_C\) and fed into the target critic \({Q^\prime}_i\) along with \(x^\prime_{LSTM}\) to produce two estimate Q-values from which the most pessimistic value (maximum) is taken in computing the time discounted reward of the Bellman equation: \({Q^\prime}_v = r + \gamma  \)Max\(({Q^\prime}_i(x^\prime_{LSTM}, [s^\prime_C, a^\prime]))\). The temporal LSTM embedding for the current state is then computed using the source Temporal-Encoder which is subsequently fed into the source critic from which the average critic Q-value error is calculated: \(Loss_c = Q_i(q_T(s_T), [s_C, a]) - {Q^\prime}_v\). Gradients are then propagated through \(Q_i\) and \(q_T\) with their respective weights updated.

For Step~3, actor optimisation, the current temporal LSTM state embedding \(x_{LSTM}\) is initially calculated using the source Temporal-Encoder \(q_T\). \(x_{LSTM}\) is then fed into the source actor along with the current actor state \(s_A\), whose output is concatenated with \(s_C\) and fed into the first source critic alongside the previously computed \(x_{LSTM}\): \(Q_v = Q_1(x_{LSTM}, [s_C, \pi(x_{LSTM}, s_A)])\). \(\pi\) and \(q_T\) are both optimised via policy gradient by minimising \(Q_v\), where caution needs to be taken to ensure that gradients are not propagated directly from the critic to the Temporal-Encoder i.e. \(Q_1  \rightarrow q_T\), but must first pass through the actor: \(Q_1 \rightarrow \pi \rightarrow q_T\) to avoid \(q_T\) directly being optimised by the critic during the actor optimisation step. The actor optimisation step is performed every second iteration.

Information regarding training of the policy module including input state information, reward structuring and hyperparameters can be found in Appendix \ref{PolicyImplementation}.

\section{Model Training Results}
\subsection{Perception Module}
The aim of introducing an identical latent representation for both high quality and simplistic images is the reduction in image rendering time during exploration. Introducing the CM-SVAE architecture allowed for a total of 24 UAVs flying simultaneously at 2x simulation speed whereas in our prior work \cite{Kal2}, a total of 16 UAVs simultaneously flying at 1x simulation speed was achievable on identical hardware. The CM-SVAE architecture accounted for a reduction in exploration time by a factor of 3, a significant improvement. When further accounting for the increase in image size from our prior work of 141w \(\times\) 80h to 210w \(\times\) 120h in our current work, the introduction of the simplistic image representation allowed for an increased rendered pixel output by a factor of 6.7, without any noticeable loss in network performance and capable of transferring to real unseen environments.

\subsection{Policy Module Ablation Study} \label{Result_TD3}
To quantify the impact of network architecture design decisions from the introduction of Shared-TD3, an ablation study was performed where three unique models were trained. (i) Shared-TD3, the proposed model architecture outlined in Section \ref{Shared-TD3 Algorithm} where the Temporal-Encoder network is shared by the actor, critic and the Temporal-Decoder. (ii) No temporal decoder, where the Temporal-Encoder is shared by both the actor and critic but the Temporal-Decoder is omitted from the model. (iii) Standard-TD3, which follows the base algorithm outlined in \cite{TD3} where both actor and critic have their own unique LSTM temporal branch similar to our prior work \cite{Kal2}. Each of the three models were trained over three random initialisations, where each initialisation consisted of four-million training iterations.

\subsubsection{Policy Module Ablation: Training Results}
\begin{figure*}
\centering
\includegraphics[width=1.0\columnwidth]{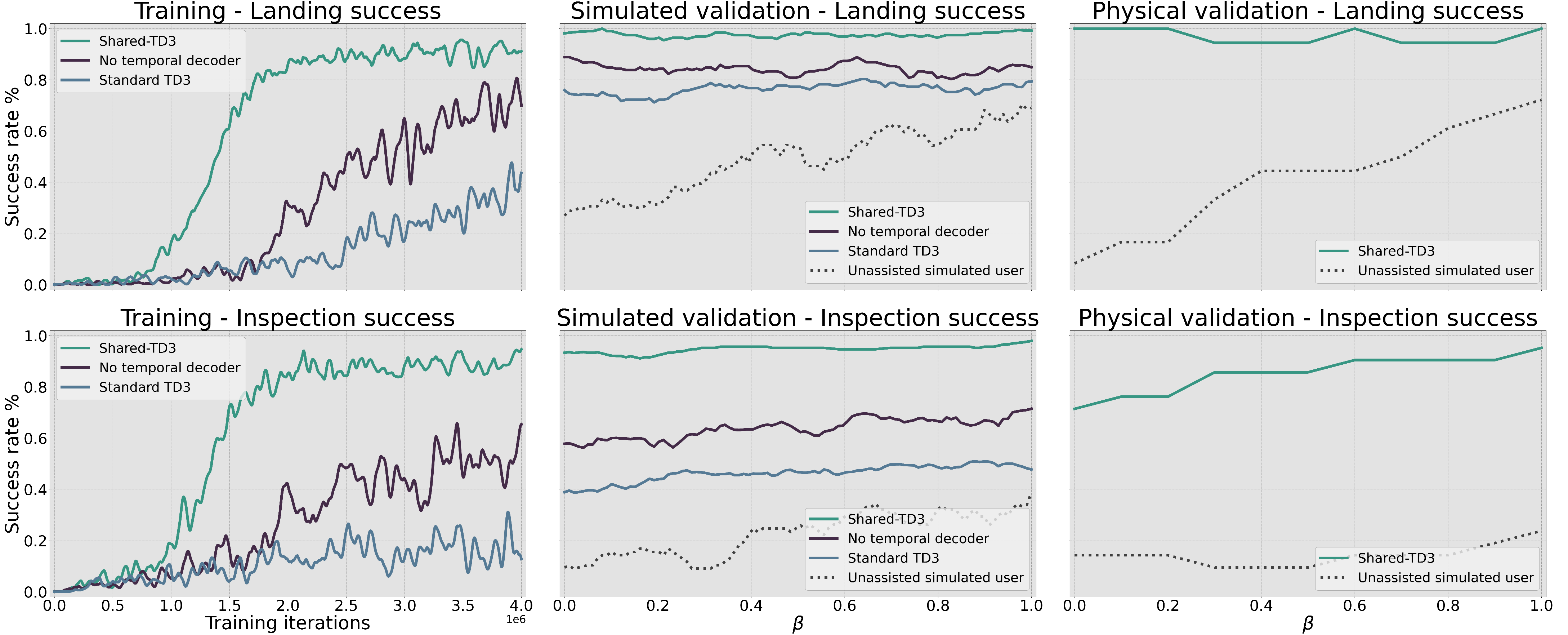}
\caption{Architecture ablation study results. Each metric is averaged over the three training initialisations. 
For the top plots, a landing is considered a success if the UAV lands on the intended platform with all four legs contacting the surface with a relative yaw error of less than 20 degrees. For the bottom plots, an inspection is considered a success if all four corners of the intended platform are within the captured image with a relative yaw error of less than 20 degrees. 
Left column: model success rates whilst training. 
Middle column: model success rates during simulated validation. 
Right column: model success rates during physical validation. 
}
\label{AblationResults}
\end{figure*}

The training results of the ablation study can be seen in the left column of Fig.~\ref{AblationResults}.
The proposed model Shared-TD3 was found to have the best performance in training with a final average success rate for the landing and inspection task of [91.7\% \& 94.6\%], compared to [70.8\% \& 65.3\%] and [41.7\% \& 12.8\%] for the no temporal decoder and standard TD3 models respectively.  

Introducing a shared Temporal-Encoder not only reduced the iterations required to learn a suitable policy, where the Shared-TD3 and no temporal decoder model was able to achieve the standard TD3 landing success rate within 33\% and 58\% of the total training iterations respectively, but also reduced the time taken to perform a single training optimisation iteration. Using a batch size of 64 and a LSTM lookback range of 50 iterations, which is equivalent to 10 seconds worth of prior state-transitions, the standard TD3 model took an average 39.75ms to perform a single optimisation iteration. Compared to the no temporal decoder model taking on average 27.95ms per iteration, 70\% of the required time for the standard TD3 algorithm. The Shared-TD3 model took on average 36.55ms per iteration, 92\% of the required time for the standard TD3 algorithm despite an additional network, the Temporal-Decoder, needing to be trained. The reduction in time per optimisation iteration is due to reusing previously computed LSTM encoding branch outputs during training, which is the most computationally inefficient component of the network forward passes. All tests were performed using an RTX-3090 GPU and AMD-5950 CPU.

\subsubsection{Policy Module Ablation: Simulated Validation}
The ablated models were subjected to a standardized validation sequence consisting of 6 missions. The simulation environment and mission plan reflected the conditions that were used in the user study outlined in Section \ref{UserStudySection}. For each model initialisation, the sequence of 6 missions were performed a total of 101 times, where \(\beta\) was swept from 0.0 to 1.0 in 0.01 increments while the remaining simulated user’s parameters \(\alpha\), \(\Psi\) and \(\Phi\) were held constant at 0.5. To ensure that the simulated user’s behaviour remained consistent throughout each \(\beta\) value for each of the model initialisations, a unique number to seed the simulated user’s random number generators was assigned to each of the 6 missions.

\subsubsection{Policy Module Ablation: Simulated Validation Results}
The simulated validation results can be seen in the middle column of Fig.~\ref{AblationResults}.
The highest performing network architecture was the proposed Shared-TD3 model with an average landing and inspection success rate of [97.6\% \& 94.6\%], followed by the no temporal decoder model at [84.4\% \& 63.9\%], then the standard TD3 model at [76.4\% \& 46.3\%]. The baseline simulated user without any additional assistance scored an average landing and inspection success rate of [48.8\% \& 22.6\%]. 

The Shared-TD3 model demonstrated an equal ability in completing both the landing and inspection task whilst the no temporal decoder model and standard TD3 model demonstrated a disparity between the landing and inspection tasks, where both models had poorer performance in the inspection task. Observing the trajectories during the inspection task, the average iterations that the assistant captures an image after the simulated user’s request for the no temporal decoder and standard TD3 model is 1.0 (0.2s) and 0.4 (.08s) respectively, compared to the Shared-TD3 model of 7.5 (1.5s). This indicates that without the Temporal-Decoder to forcefully condition the Temporal-Encoder to extract temporal information regarding the simulated user’s current image request status, the models have difficulties to extract this information via reinforcement learning and instead request an image whilst the simulated user’s image request is directly observable within the current input. This concept is further strengthened by observing the ratio of total image requests of the assistant divided by the image requests of the simulated user for the no temporal decoder and standard TD3 models, which have a value of 2.22 and 3.25 respectively compared to a value of 1.01 for the Shared-TD3 model. This indicates that the two models without the Temporal-Decoder struggle to extract the temporal information required to remember whether they have already taken a picture for the given task. 

\subsubsection{Policy Module Ablation: Physical Validation}
To assess the performance of the proposed Shared-TD3 model when transferring from simulation to reality, an identical set of 6 missions performed in the simulated validation were performed in a physical environment where the simulated user piloted the physical UAV. 

\subsubsection{Policy Module Ablation: Physical Validation Results}
The results of the physical validation can be seen in the right column of Fig.~\ref{AblationResults}.
The performance of the Shared-TD3 model was comparable to that of the simulated validation experiment where only a slightly lower success rate for the inspection task was observed for lower \(\beta\) simulated users. The Shared-TD3 model was successfully able to be transferred to reality despite being trained only in simulation, on simplistic input images.

\section{Supplementary Information Schemes}\label{SupInformationSection}

To further enhance the shared autonomy assistant discussed in Section \ref{PolicyModelSection}, supplementary information schemes are proposed to provide pilots with a greater understanding of the current state of the UAV and the assistant. From our prior work \cite{Kal2}, participants stated that they experienced the greatest difficulties with estimating the depth of the UAV, resulting in failed landings primarily from position errors along the depth axis. The difficulty to correctly infer the depth of the UAV motivates our approach to supply supplementary information to aid pilots in perceiving the current spatial state of the UAV. Further issues pertaining to a lack of information arose from participants stating that their most disliked aspect of the assistant was the inability to perceive its intent. The lack of communication of the assistant left participants with a sense of uncertainty in what was expected of them and their role during the mission. Therefore supplementary information schemes for shared autonomy UAV missions should focus on two aspects: (i) providing spatial information of the UAV relative to the environment and (ii) providing information about the current intent of the assistant and what the assistant expects of the pilot.

\begin{figure}[t]
\centering
\includegraphics[width=1.0\columnwidth]{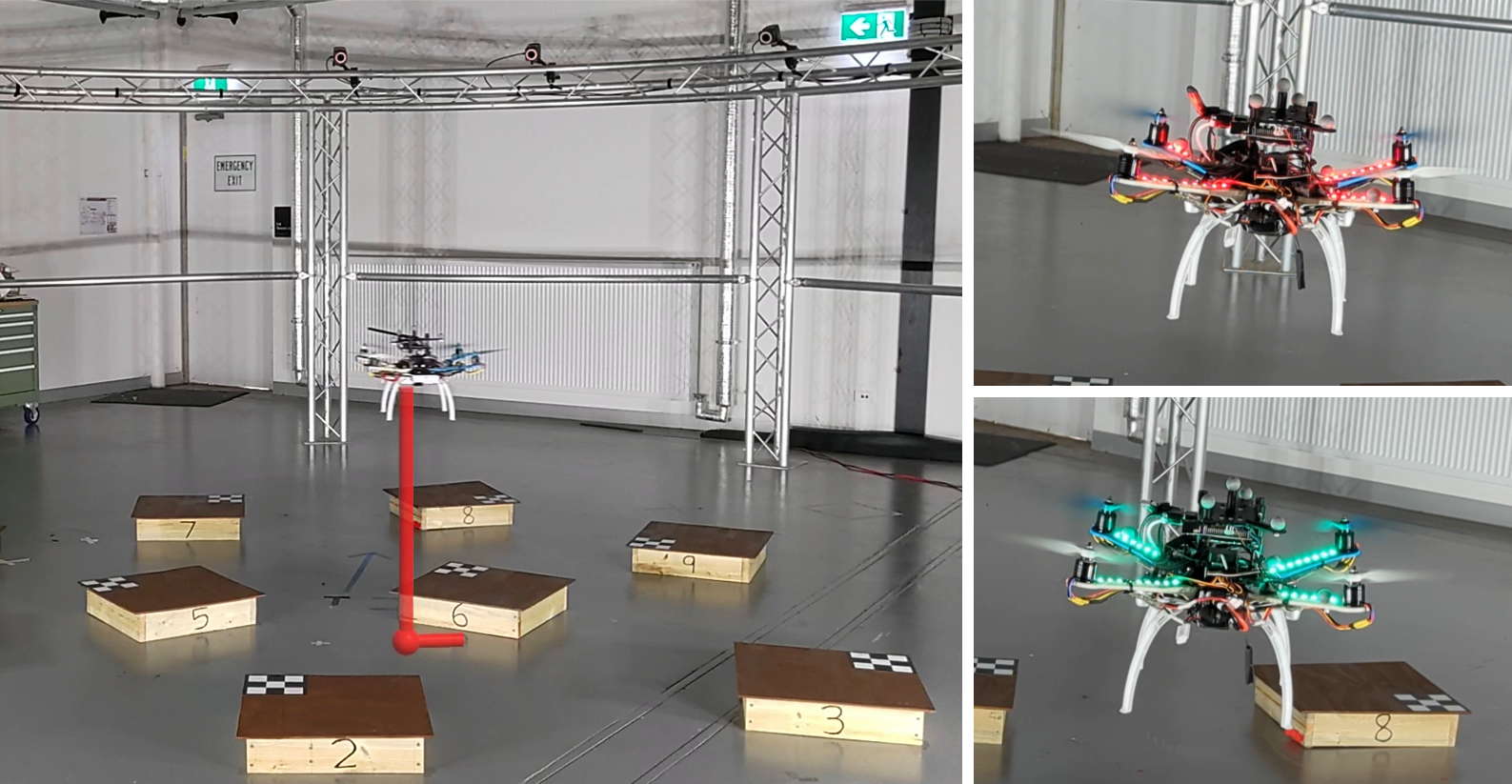}
\caption{(Left) Example of the augmented reality display from the HoloLens. (Right) Example of red and green light feedback.}
\label{HoloLensLightExample}
\end{figure}

When developing an information display interface for UAV applications, maintaining eye-contact with the UAV is important to prevent mental burden and lapses of concentration from context switching between the information interface and the UAV. Therefore proposed information displays should either be contained onboard the UAV, or augment the pilot’s natural vision with augmented reality displays. For providing information about the state/intent of the assistant, onboard information displays are chosen using onboard LED lights to display red or green lights. Red/green lights are chosen for an intuitive, universally understood representation of being in a bad/good condition which the assistant learns to display to promote specific behaviours as discussed in Section \ref{PolicyModelSimUser}. To convey spatial information of the UAV relative to the environment an augmented reality display is chosen using the HoloLens. The augmented reality display implemented in the HoloLens projects a vertical line from the centroid of the drone onto the floor to help pilots visualise the depth of the UAV. A secondary horizontal line whose endpoint is coincident to the vertical line's endpoint on the floor is rendered to display the UAV's current orientation. An example of the HoloLens augmented reality display and red/green light display can be seen in Fig.~\ref{HoloLensLightExample}.

\section{User Study}\label{UserStudySection}
To assess the performance of the proposed assistant and to determine what preferences human pilots have in regards to what additional information is provided to them, a user study was conducted. The user study follows a within-subject design which was approved by the Monash University Human Research Ethics Committee (MUHREC), project ID 35281.

\subsection{Task and Environment}
\begin{figure}
\centering
\includegraphics[width=1.0\columnwidth]{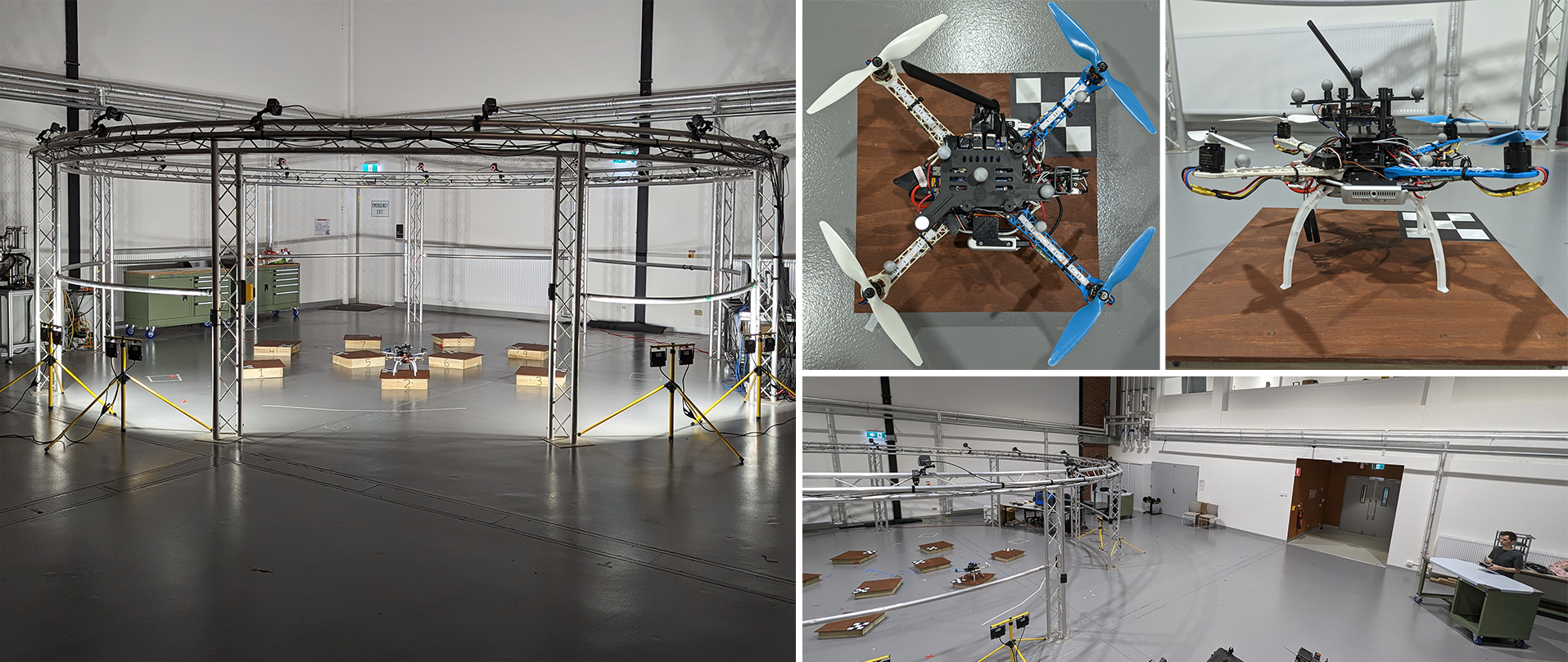}
\caption{Physical layout of user study environment. (Left) View of the arena from participant’s perspective. (Top right) UAV used in user study resting atop of landing platform. (Bottom right) Participants’ relative position to the arena.}
\label{PhysicalArena}
\end{figure}

Participants were tasked with completing multi-task missions comprised of both inspection and landing tasks.
The physical arena in which participants flew the drone consisted of nine labeled platforms of size 0.5\(\times\)0.5\(\times\)0.12m, which were placed within a 7.2m diameter circular truss. 
Participants completed the study under a total of 3 conditions, where 
each condition contained an identical set of 6 missions. Each mission comprised of \(0- 2\) inspection tasks followed by a landing task. Each condition consisted of seven inspection tasks and six landing tasks, for a total of 21 inspections and 18 landings for the entire study per participant. 
During each mission participants stood 11m away from the center of the arena to replicate our prior study's conditions \cite{Kal2}, where this distance was found to be sufficient in introducing depth perception issues that differentiate expert pilots from novices. Participants were instructed to remain stationary throughout each mission. The physical layout of the user study arena can be seen in Fig.~\ref{PhysicalArena}.

\subsection{Study Conditions}
The three conditions participants flew the drone under were Assisted, Assisted-Lights and Assisted-HoloLens. For the Assisted condition, participants flew the drone using the proposed model outlined in Section \ref{PolicyModelSection} but where the assistant's red and green light display action was hidden to the pilot. For the Assisted-Lights condition, participants flew with an identical assistant to that of the Assisted condition however the assistant's red and green light display action was shown to participants. For the Assisted-HoloLens condition, participants flew with an identical assistant to that of the Assisted condition where the assistant's red and green light display action remained hidden to the pilot but instead were asked to fly wearing the HoloLens augmented reality display as disscussed in Section \ref{SupInformationSection}. For each condition, participants were not told the specific details of the assistant and how it operates but instead had to figure out themselves the optimal strategy for working with it. Participants were not explicitly told the meaning behind the red/green lights but were instead prompted to think about what a red or green light could indicate.
Omitting operational details about the assistant is to test the intuitiveness of the assistant, ensuring that novice users without any form of training can work alongside the assistant. Assistant details are also omitted to prevent biasing participant behavior towards a policy which may be optimal for the assistant. Instead participants are encouraged to adopt a policy in which they prefer to test the generality of the assistant.

\subsection{Assessment metrics}
Participants were assessed under five metrics: success rate, position error, yaw error, time / initial distance and distance traveled / initial distance. Success for the landing task is defined as the UAV remaining at rest a top of the correct platform with a yaw error less than 20 degrees. Success for the inspection task is defined as having all four corners of the platform visible within the downwards facing cameras with a yaw error less than 20 degrees. Position error for landing and inspection tasks is defined as the XY distance between the platform’s centroid and the drone’s landing position or image capture position. Yaw error is defined as the difference of orientation along the z-axis between the platform and the drone’s landing or image capture orientation. Both position and yaw error aim to measure task accuracy. Time / initial distance and distance traveled / initial distance is the time or distance traveled to complete a task, normalized by the initial XY distance from where the task starts to the intended goal location. The time and distance traveled metrics are normalized by the initial starting distance so that tasks within a mission can be compared to tasks across different missions as tasks with larger initial starting distances require the drone to travel more and hence take more time to complete. Both time and distance traveled metrics aim to measure task completion efficiency. All metrics derive from continuous variables excluding the success rate which derives from binary count data.

\subsection{UAV Hardware}
Participants flew a custom-built UAV using the Pixhawk Cube flight controller running ArduPilot. An Odroid-N2+ was used as an onboard companion computer to poll RGB-D images from two Intel RealSense D435i cameras situated on the left and right of the UAV. All perception and policy module network inference was performed using the onboard Odroid companion computer. Fourteen Bonita-10 Vicon motion capture cameras were used for recording UAV pose information which were wirelessly sent to the Odroid, then subsequently forwarded onto the flight controller to allow for pose stabilization in the GPS denied environment. Participants were constrained to fly within a 6m safe-to-fly zone by an automatic safety system. The safety system alerted participants by strobing RGB LED light strips attached to the arms of the drone and would forcibly take over and push the drone towards the center of the arena if the pilot’s actions were deemed unsafe. 

\subsection{Study Procedures}
Participants were initially asked to fill out a demographics survey inquiring about their previous drone, joystick and video game experience as well as if they participated in our previous study \cite{Kal2}. Participants were then shown an introductory video explaining the controls of the UAV and the general user study procedure. Participants were then given the opportunity to practice flying the UAV to understand the control scheme and dynamics of the drone. After the conclusion of the practice flight, the ceiling lights were turned off and eight LED flood lights situated around the base of the arena were turned on to mimic the user study conditions in \cite{Kal2} so that participants could not rely on well-defined shadows cast by the UAV as a substitute for depth perception. 

Participants were then initially assigned a condition out of the three potential conditions based on their participant ID number. To minimize the impact of learning biases from condition ordering, the order in which each participant completed the 3 conditions was determined by selecting one of the 6 possible ordering permutations using their participant ID as an index. Participants received a mission plan detailing the required tasks to perform for each of the six missions within the condition. To aid the participant’s memory, the experimenter briefly shone a laser light at the center of the next goal platform after the completion of each task within the mission. The first author initiated the take-off and brought the UAV to an altitude of 1.4m using a separate master RC transmitter, after which the LED lights on the UAV strobed to indicate the participant can start the mission. 

After completing all six missions within the condition, participants were then given a NASA Task Load Index (TLX) \cite{TLX} survey as well as 5-point scale multiple choice questions querying how they perceived the given condition. Participants then repeated the process of completing the remaining two conditions and their associated surveys, the order of which determined by their participant ID number. After completing all three conditions, participants were then given the final survey which consisted of several short answer questions. The entire user study took on average 1.5 hours for each participant to complete. The full list of survey questions can be seen in Table~\ref{DemographicsTable}, Table~\ref{ConditionPerceptionTable} \& Table~\ref{WordedResponse}, alongside the user study results in Section \ref{ResultsSection}.

\subsection{Sub-User Study}
An additional sub-user study was held in the subsequent week after the conclusion of the main user study. Participants from the main user study returned to complete six missions which identically mimicked the six missions from the main user study but without any assistance from the assistant, light feedback or HoloLens. Participants took on average 20 minutes to complete the sub-user study and were required to complete the TLX survey for the unassisted condition. The sub-user study’s aim was to reaffirm the findings of our prior work \cite{Kal2}, that an assistant is necessary to ensure task success, compared to evaluating pilot preference and performance towards auxiliary information for the main study. The unassisted missions in the sub-user study were held after the main user study due to participant time constraints and potential fatigue induced from extended periods of flying. The sub-user study was approved under the same MUHREC project ID 35281.

\section{User Study Results}\label{ResultsSection}
A total of 29 participants completed the main user study for a total of 174 landings and 203 inspections for each of the three assisted conditions. For the sub-user study a total of 10 participants returned to complete a total of 60 landings and 70 inspections in the unassisted condition. Participant demographic information can be found in Table~\ref{DemographicsTable}.

\newcommand{\DrawCell}[4] 
  {
  \def\TL{#1}
  \def\TR{#2}
  \def\BR{#3}
  \def\BL{#4}
  \draw (\TL) -- (\TR) -- (\BR) -- (\BL) -- (\TL);
  }
  
\newcommand{\DrawBottomlessCell}[4]
  {
  \def\TL{#1}
  \def\TR{#2}
  \def\BR{#3}
  \def\BL{#4}
  \draw (\BL)  -- (\TL) -- (\TR) -- (\BR);
  }
  
\newcommand{\DrawToplessCell}[4] 
  {
  \def\TL{#1}
  \def\TR{#2}
  \def\BR{#3}
  \def\BL{#4}
  \draw (\TL)  -- (\BL) -- (\BR) -- (\TR);
  }

\tikzset{ 
    DemographicSurveyTable/.style={
        matrix of nodes,
        nodes={
            rectangle,
            draw=white,
            align=center,
            text width=14.0mm
        },
        minimum height=3.0mm,
        text depth=2.5mm,
        text height=2.5mm,
        font=\footnotesize,
        nodes in empty cells,
        align=center,
        column 1/.style={
            nodes={text width=30mm, align=left}
        }
    }
}

\tikzset{ 
    ConditionPerceptionTable/.style={
        matrix of nodes,
        nodes={
            rectangle,
            draw=white,
            align=center,
        },
        minimum height=3.0mm,
        text depth=2.5mm,
        text height=1.5mm,
        font=\footnotesize,
        nodes in empty cells,
        align=center,
        column 1/.style={
            nodes={text width=60.0mm, align=justify}
        },
        row 1/.style={
            nodes={
                font=\bfseries, align=center, text depth=0.5mm, text height=2mm, minimum height=2mm
            }
        }
    }
}

\begin{figure}[!b]
\captionof{table}{Participant Demographics}
\label{DemographicsTable}

\begin{tikzpicture}
 \matrix[DemographicSurveyTable] (Demo)
 {
&&&&&\\
&&&&&\\
&&&&&\\
&&&&&\\
&&&&&\\
&&&&&\\
&&&&&\\
&&&&&\\
};
\definecolor{OddColor}{rgb}{0.3333333333333333, 0.47843137254901963, 0.5843137254901961}
\definecolor{EvenColor}{rgb}{0.21568627450980393, 0.5882352941176471, 0.5137254901960784}

\DrawBottomlessCell{Demo-1-1.north west}{Demo-1-1.north east}{Demo-1-1.south east}{Demo-1-1.south west}
\DrawToplessCell{Demo-2-1.north west}{Demo-2-1.north east}{Demo-2-1.south east}{Demo-2-1.south west}
\node[anchor = north west, align=justify, text width=30mm, font=\footnotesize] at (Demo-1-1.north west) {Did you participate in our previous physical drone landing user study?};
\DrawCell{Demo-1-2.north west}{Demo-1-4.north}{Demo-1-4.south}{Demo-1-2.south west}
\DrawCell{Demo-1-4.north}{Demo-1-6.north east}{Demo-1-6.south east}{Demo-1-4.south}
\DrawCell{Demo-2-2.north west}{Demo-2-4.north}{Demo-2-4.south}{Demo-2-2.south west}
\DrawCell{Demo-2-4.north}{Demo-2-6.north east}{Demo-2-6.south east}{Demo-2-4.south}
\node[anchor = north, align=justify, text width=14.0mm, font=\footnotesize] at (Demo-1-3.north) {\\ No};
\node[anchor = north, align=justify, text width=14.0mm, font=\footnotesize] at (Demo-1-5.north east) {\\ Yes};
\node[anchor = north, align=justify, text width=14.0mm, font=\footnotesize] at (Demo-2-3.north) {\\ \textbf{62\%}};
\node[anchor = north, align=justify, text width=14.0mm, font=\footnotesize] at (Demo-2-5.north east) {\\ \textbf{37\%}};
\draw[opacity=0.6,fill=OddColor] ($ (Demo-2-4.north)!1 - 62 / 100!(Demo-2-4.south) $) rectangle(Demo-2-2.south west);
\draw[opacity=0.6,fill=OddColor] ($ (Demo-2-6.north east)!1 - 37 / 100!(Demo-2-6.south east) $) rectangle(Demo-2-4.south);

\DrawBottomlessCell{Demo-3-1.north west}{Demo-3-1.north east}{Demo-3-1.south east}{Demo-3-1.south west}
\DrawToplessCell{Demo-4-1.north west}{Demo-4-1.north east}{Demo-4-1.south east}{Demo-4-1.south west}
\node[anchor = north west, align=justify, text width=30mm, font=\footnotesize] at (Demo-3-1.north west) {How many hours of experience have you with flying physical drones?};
\DrawCell{Demo-3-2.north west}{Demo-3-2.north east}{Demo-3-2.south east}{Demo-3-2.south west}
\DrawCell{Demo-4-2.north west}{Demo-4-2.north east}{Demo-4-2.south east}{Demo-4-2.south west}
\node[anchor = north west, align=center, text width=14.0mm, font=\footnotesize] at (Demo-3-2.north west) {\\ 0-5};
\node[anchor = north west, align=center, text width=14.0mm, font=\footnotesize] at (Demo-4-2.north west) {\\ \textbf{68\%}};
\draw[opacity=0.6,fill=EvenColor] ($ (Demo-4-2.north west)!1 - 68 / 100!(Demo-4-2.south west) $) rectangle(Demo-4-2.south east);
\DrawCell{Demo-3-3.north west}{Demo-3-3.north east}{Demo-3-3.south east}{Demo-3-3.south west}
\DrawCell{Demo-4-3.north west}{Demo-4-3.north east}{Demo-4-3.south east}{Demo-4-3.south west}
\node[anchor = north west, align=center, text width=14.0mm, font=\footnotesize] at (Demo-3-3.north west) {\\ 5-10};
\node[anchor = north west, align=center, text width=14.0mm, font=\footnotesize] at (Demo-4-3.north west) {\\ \textbf{13\%}};
\draw[opacity=0.6,fill=EvenColor] ($ (Demo-4-3.north west)!1 - 13 / 100!(Demo-4-3.south west) $) rectangle(Demo-4-3.south east);
\DrawCell{Demo-3-4.north west}{Demo-3-4.north east}{Demo-3-4.south east}{Demo-3-4.south west}
\DrawCell{Demo-4-4.north west}{Demo-4-4.north east}{Demo-4-4.south east}{Demo-4-4.south west}
\node[anchor = north west, align=center, text width=14.0mm, font=\footnotesize] at (Demo-3-4.north west) {\\ 10-20};
\node[anchor = north west, align=center, text width=14.0mm, font=\footnotesize] at (Demo-4-4.north west) {\\ \textbf{6\%}};
\draw[opacity=0.6,fill=EvenColor] ($ (Demo-4-4.north west)!1 - 6 / 100!(Demo-4-4.south west) $) rectangle(Demo-4-4.south east);
\DrawCell{Demo-3-5.north west}{Demo-3-5.north east}{Demo-3-5.south east}{Demo-3-5.south west}
\DrawCell{Demo-4-5.north west}{Demo-4-5.north east}{Demo-4-5.south east}{Demo-4-5.south west}
\node[anchor = north west, align=center, text width=14.0mm, font=\footnotesize] at (Demo-3-5.north west) {\\ 20-50};
\node[anchor = north west, align=center, text width=14.0mm, font=\footnotesize] at (Demo-4-5.north west) {\\ \textbf{3\%}};
\draw[opacity=0.6,fill=EvenColor] ($ (Demo-4-5.north west)!1 - 3 / 100!(Demo-4-5.south west) $) rectangle(Demo-4-5.south east);
\DrawCell{Demo-3-6.north west}{Demo-3-6.north east}{Demo-3-6.south east}{Demo-3-6.south west}
\DrawCell{Demo-4-6.north west}{Demo-4-6.north east}{Demo-4-6.south east}{Demo-4-6.south west}
\node[anchor = north west, align=center, text width=14.0mm, font=\footnotesize] at (Demo-3-6.north west) {\\ 50+};
\node[anchor = north west, align=center, text width=14.0mm, font=\footnotesize] at (Demo-4-6.north west) {\\ \textbf{6\%}};
\draw[opacity=0.6,fill=EvenColor] ($ (Demo-4-6.north west)!1 - 6 / 100!(Demo-4-6.south west) $) rectangle(Demo-4-6.south east);

\DrawBottomlessCell{Demo-5-1.north west}{Demo-5-1.north east}{Demo-5-1.south east}{Demo-5-1.south west}
\DrawToplessCell{Demo-6-1.north west}{Demo-6-1.north east}{Demo-6-1.south east}{Demo-6-1.south west}
\node[anchor = north west, align=justify, text width=30mm, font=\footnotesize] at (Demo-5-1.north west) {How often do you play video games?};
\DrawCell{Demo-5-2.north west}{Demo-5-2.north east}{Demo-5-2.south east}{Demo-5-2.south west}
\DrawCell{Demo-6-2.north west}{Demo-6-2.north east}{Demo-6-2.south east}{Demo-6-2.south west}
\node[anchor = north west, align=center, text width=14.0mm, font=\footnotesize] at (Demo-5-2.north west) {\\ Never};
\node[anchor = north west, align=center, text width=14.0mm, font=\footnotesize] at (Demo-6-2.north west) {\\ \textbf{24\%}};
\draw[opacity=0.6,fill=OddColor] ($ (Demo-6-2.north west)!1 - 24 / 100!(Demo-6-2.south west) $) rectangle(Demo-6-2.south east);
\DrawCell{Demo-5-3.north west}{Demo-5-3.north east}{Demo-5-3.south east}{Demo-5-3.south west}
\DrawCell{Demo-6-3.north west}{Demo-6-3.north east}{Demo-6-3.south east}{Demo-6-3.south west}
\node[anchor = north west, align=center, text width=14.0mm, font=\footnotesize] at (Demo-5-3.north west) {\\ Monthly};
\node[anchor = north west, align=center, text width=14.0mm, font=\footnotesize] at (Demo-6-3.north west) {\\ \textbf{34\%}};
\draw[opacity=0.6,fill=OddColor] ($ (Demo-6-3.north west)!1 - 34 / 100!(Demo-6-3.south west) $) rectangle(Demo-6-3.south east);
\DrawCell{Demo-5-4.north west}{Demo-5-4.north east}{Demo-5-4.south east}{Demo-5-4.south west}
\DrawCell{Demo-6-4.north west}{Demo-6-4.north east}{Demo-6-4.south east}{Demo-6-4.south west}
\node[anchor = north west, align=center, text width=14.0mm, font=\footnotesize] at (Demo-5-4.north west) {\\ Weekly};
\node[anchor = north west, align=center, text width=14.0mm, font=\footnotesize] at (Demo-6-4.north west) {\\ \textbf{20\%}};
\draw[opacity=0.6,fill=OddColor] ($ (Demo-6-4.north west)!1 - 20 / 100!(Demo-6-4.south west) $) rectangle(Demo-6-4.south east);
\DrawCell{Demo-5-5.north west}{Demo-5-5.north east}{Demo-5-5.south east}{Demo-5-5.south west}
\DrawCell{Demo-6-5.north west}{Demo-6-5.north east}{Demo-6-5.south east}{Demo-6-5.south west}
\node[anchor = north west, align=center, text width=14.0mm, font=\footnotesize] at (Demo-5-5.north west) {\\ Regularly};
\node[anchor = north west, align=center, text width=14.0mm, font=\footnotesize] at (Demo-6-5.north west) {\\ \textbf{17\%}};
\draw[opacity=0.6,fill=OddColor] ($ (Demo-6-5.north west)!1 - 17 / 100!(Demo-6-5.south west) $) rectangle(Demo-6-5.south east);
\DrawCell{Demo-5-6.north west}{Demo-5-6.north east}{Demo-5-6.south east}{Demo-5-6.south west}
\DrawCell{Demo-6-6.north west}{Demo-6-6.north east}{Demo-6-6.south east}{Demo-6-6.south west}
\node[anchor = north west, align=center, text width=14.0mm, font=\footnotesize] at (Demo-5-6.north west) {\\ Daily};
\node[anchor = north west, align=center, text width=14.0mm, font=\footnotesize] at (Demo-6-6.north west) {\\ \textbf{3\%}};
\draw[opacity=0.6,fill=OddColor] ($ (Demo-6-6.north west)!1 - 3 / 100!(Demo-6-6.south west) $) rectangle(Demo-6-6.south east);

\DrawBottomlessCell{Demo-7-1.north west}{Demo-7-1.north east}{Demo-7-1.south east}{Demo-7-1.south west}
\DrawToplessCell{Demo-8-1.north west}{Demo-8-1.north east}{Demo-8-1.south east}{Demo-8-1.south west}
\node[anchor = north west, align=justify, text width=30mm, font=\footnotesize] at (Demo-7-1.north west) {How many hours have you spent using a joystick controller?};
\DrawCell{Demo-7-2.north west}{Demo-7-2.north east}{Demo-7-2.south east}{Demo-7-2.south west}
\DrawCell{Demo-8-2.north west}{Demo-8-2.north east}{Demo-8-2.south east}{Demo-8-2.south west}
\node[anchor = north west, align=center, text width=14.0mm, font=\footnotesize] at (Demo-7-2.north west) {\\ 0-10};
\node[anchor = north west, align=center, text width=14.0mm, font=\footnotesize] at (Demo-8-2.north west) {\\ \textbf{31\%}};
\draw[opacity=0.6,fill=EvenColor] ($ (Demo-8-2.north west)!1 - 31 / 100!(Demo-8-2.south west) $) rectangle(Demo-8-2.south east);
\DrawCell{Demo-7-3.north west}{Demo-7-3.north east}{Demo-7-3.south east}{Demo-7-3.south west}
\DrawCell{Demo-8-3.north west}{Demo-8-3.north east}{Demo-8-3.south east}{Demo-8-3.south west}
\node[anchor = north west, align=center, text width=14.0mm, font=\footnotesize] at (Demo-7-3.north west) {\\ 10-25};
\node[anchor = north west, align=center, text width=14.0mm, font=\footnotesize] at (Demo-8-3.north west) {\\ \textbf{10\%}};
\draw[opacity=0.6,fill=EvenColor] ($ (Demo-8-3.north west)!1 - 10 / 100!(Demo-8-3.south west) $) rectangle(Demo-8-3.south east);
\DrawCell{Demo-7-4.north west}{Demo-7-4.north east}{Demo-7-4.south east}{Demo-7-4.south west}
\DrawCell{Demo-8-4.north west}{Demo-8-4.north east}{Demo-8-4.south east}{Demo-8-4.south west}
\node[anchor = north west, align=center, text width=14.0mm, font=\footnotesize] at (Demo-7-4.north west) {\\ 25-100};
\node[anchor = north west, align=center, text width=14.0mm, font=\footnotesize] at (Demo-8-4.north west) {\\ \textbf{10\%}};
\draw[opacity=0.6,fill=EvenColor] ($ (Demo-8-4.north west)!1 - 10 / 100!(Demo-8-4.south west) $) rectangle(Demo-8-4.south east);
\DrawCell{Demo-7-5.north west}{Demo-7-5.north east}{Demo-7-5.south east}{Demo-7-5.south west}
\DrawCell{Demo-8-5.north west}{Demo-8-5.north east}{Demo-8-5.south east}{Demo-8-5.south west}
\node[anchor = north west, align=center, text width=14.0mm, font=\footnotesize] at (Demo-7-5.north west) {\\ 100-200};
\node[anchor = north west, align=center, text width=14.0mm, font=\footnotesize] at (Demo-8-5.north west) {\\ \textbf{17\%}};
\draw[opacity=0.6,fill=EvenColor] ($ (Demo-8-5.north west)!1 - 17 / 100!(Demo-8-5.south west) $) rectangle(Demo-8-5.south east);
\DrawCell{Demo-7-6.north west}{Demo-7-6.north east}{Demo-7-6.south east}{Demo-7-6.south west}
\DrawCell{Demo-8-6.north west}{Demo-8-6.north east}{Demo-8-6.south east}{Demo-8-6.south west}
\node[anchor = north west, align=center, text width=14.0mm, font=\footnotesize] at (Demo-7-6.north west) {\\ 200+};
\node[anchor = north west, align=center, text width=14.0mm, font=\footnotesize] at (Demo-8-6.north west) {\\ \textbf{31\%}};
\draw[opacity=0.6,fill=EvenColor] ($ (Demo-8-6.north west)!1 - 31 / 100!(Demo-8-6.south west) $) rectangle(Demo-8-6.south east);

\end{tikzpicture}
 \end{figure}

\subsection{Task Performance Results}\label{UserStudyResultsQuantitative}
A summary of key performance metrics can be viewed in Table~\ref{UserStudySummaryTable}.

\definecolor{TableGray}{gray}{0.9}
\newcommand\SummaryTableFirstWidth{3.0}
\newcommand\SumaryTableSecondWidth{1.6}
\newcommand\SummaryTableWidth{2.8}
\newcommand\SummaryTableHSpace{0.125}
\newcolumntype{P}[1]{>{\centering\arraybackslash}p{#1}}

\begin{table}[t]
\begin{center}
\captionof{table}{User study performance metrics summary}
\setlength{\tabcolsep}{0.2em}
\label{UserStudySummaryTable}
\footnotesize
\begin{tabular}{|p{\SummaryTableFirstWidth cm}|c|c|c|c|}
 \hline
 \multicolumn{1}{|c|}{} & \multicolumn{1}{|P{\SumaryTableSecondWidth cm}|}{Assisted} & \multicolumn{1}{|P{\SummaryTableWidth cm}|}{Assisted + Lights} & \multicolumn{1}{|P{\SummaryTableWidth cm}|}{Assisted + HoloLens} & \multicolumn{1}{|P{\SummaryTableWidth cm}|}{Unassisted} \\
 \hline
  \multicolumn{5}{|c|}{Target approach} \\ 
  \hline
   \multicolumn{1}{|p{\SummaryTableFirstWidth cm}|}{\multirow{2}{*}{}} & \multicolumn{1}{|c|}{5.52s/m} & \multicolumn{1}{|c|}{5.67s/m} & \multicolumn{1}{|c|}{6.13s/m} & \multicolumn{1}{|c|}{5.75s/m} \\
  \multicolumn{1}{|p{\SummaryTableFirstWidth cm}|}{\multirow{-2}{*}{*Time / init. dist.}} & \multicolumn{1}{|c|}{\(\sigma\)=3.92} & \multicolumn{1}{|c|}{\(\sigma\)=3.96, t=0.538, p=.591} & \multicolumn{1}{|c|}{\(\sigma\)=5.12, t=1.843, p=0.066} & \multicolumn{1}{|c|}{\(\sigma\)=4.04, t=0.185, p=.854}\\
  \rowcolor{TableGray} \multicolumn{1}{|p{\SummaryTableFirstWidth cm}|}{\multirow{2}{*}{}} & \multicolumn{1}{|c|}{1.29} & \multicolumn{1}{|c|}{1.32} & \multicolumn{1}{|c|}{1.44} & \multicolumn{1}{|c|}{\textbf{1.09}} \\
  \rowcolor{TableGray} \multicolumn{1}{|p{\SummaryTableFirstWidth cm}|}{\multirow{-2}{*}{*Dist. traveled / init. dist.}} & \multicolumn{1}{|c|}{\(\sigma\)=0.81} & \multicolumn{1}{|c|}{\(\sigma\)=0.89, t=0.477, p=.634} & \multicolumn{1}{|c|}{\(\sigma\)=1.10, t=2.033, p=0.043} & \multicolumn{1}{|c|}{\(\sigma\)=0.55, t=3.310, p=.001}\\
  \hline
  \multicolumn{5}{|c|}{Landing task} \\ 
  \hline
  \rowcolor{TableGray} \multicolumn{1}{|p{\SummaryTableFirstWidth cm}|}{\multirow{2}{*}{}} & \multicolumn{1}{|c|}{95.98\%} & \multicolumn{1}{|c|}{96.55\%} & \multicolumn{1}{|c|}{94.25\%} & \multicolumn{1}{|c|}{\textbf{16.67\%}}\\
  \rowcolor{TableGray} \multicolumn{1}{|p{\SummaryTableFirstWidth cm}|}{\multirow{-2}{*}{Success rate}} & \multicolumn{1}{|c|}{\(\sigma\)=0.20} & \multicolumn{1}{|c|}{\(\sigma\)=0.18, \(\chi^2\)=0.091, p=.763} & \multicolumn{1}{|c|}{\(\sigma\)=0.23, \(\chi^2\)=0.600, p=.439} & \multicolumn{1}{|c|}{\(\sigma\)=0.37, \(\chi^2\)=48.00, p<.001}\\
  \multicolumn{1}{|p{\SummaryTableFirstWidth cm}|}{\multirow{2}{*}{}} & \multicolumn{1}{|c|}{0.08m} & \multicolumn{1}{|c|}{0.08m} & \multicolumn{1}{|c|}{0.08m} & \multicolumn{1}{|c|}{\textbf{0.48m}}\\
  \multicolumn{1}{|p{\SummaryTableFirstWidth cm}|}{\multirow{-2}{*}{Position} error} & \multicolumn{1}{|c|}{\(\sigma\)=0.10} & \multicolumn{1}{|c|}{\(\sigma\)=0.09, t=0.021, p=.983} & \multicolumn{1}{|c|}{\(\sigma\)=0.11, t=0.641, p=.522} & \multicolumn{1}{|c|}{\(\sigma\)=0.26, t=12.09, p<.001}\\
  \rowcolor{TableGray} \multicolumn{1}{|p{\SummaryTableFirstWidth cm}|}{\multirow{2}{*}{}} & \multicolumn{1}{|c|}{5.93 deg} & \multicolumn{1}{|c|}{4.95 deg} & \multicolumn{1}{|c|}{5.52 deg} & \multicolumn{1}{|c|}{\textbf{35.95 deg}} \\
  \rowcolor{TableGray} \multicolumn{1}{|p{\SummaryTableFirstWidth cm}|}{\multirow{-2}{*}{Yaw error}} & \multicolumn{1}{|c|}{\(\sigma\)=5.87} & \multicolumn{1}{|c|}{\(\sigma\)=4.71, t=2.417, p=.017} & \multicolumn{1}{|c|}{\(\sigma\)=6.79, t=0.725, p=.469} & \multicolumn{1}{|c|}{\(\sigma\)=39.91, t=5.924, p<.001}\\
  \multicolumn{1}{|p{\SummaryTableFirstWidth cm}|}{\multirow{2}{*}{}} & \multicolumn{1}{|c|}{13.83s/m} & \multicolumn{1}{|c|}{14.34s/m} & \multicolumn{1}{|c|}{15.73s/m} & \multicolumn{1}{|c|}{\textbf{24.53s/m}} \\
  \multicolumn{1}{|p{\SummaryTableFirstWidth cm}|}{\multirow{-2}{*}{*Time / init. dist.}} & \multicolumn{1}{|c|}{\(\sigma\)=7.03} & \multicolumn{1}{|c|}{\(\sigma\)=10.82, t=0.584, p=.560} & \multicolumn{1}{|c|}{\(\sigma\)=9.15, t=2.314, p=.022} & \multicolumn{1}{|c|}{\(\sigma\)=18.42, t=4.162, p<.001}\\
  \rowcolor{TableGray} \multicolumn{1}{|p{\SummaryTableFirstWidth cm}|}{\multirow{2}{*}{}} & \multicolumn{1}{|c|}{2.18} & \multicolumn{1}{|c|}{2.33} & \multicolumn{1}{|c|}{2.53} & \multicolumn{1}{|c|}{\textbf{3.18}} \\
  \rowcolor{TableGray} \multicolumn{1}{|p{\SummaryTableFirstWidth cm}|}{\multirow{-2}{*}{*Dist. traveled / init. dist.}} & \multicolumn{1}{|c|}{\(\sigma\)=1.22} & \multicolumn{1}{|c|}{\(\sigma\)=1.72, t=0.982, p=.328} & \multicolumn{1}{|c|}{\(\sigma\)=1.62, t=2.440, p=.016} & \multicolumn{1}{|c|}{\(\sigma\)=2.33, t=3.248, p=.002}\\
 \hline
   \multicolumn{5}{|c|}{Inspection task} \\ 
  \hline
  \rowcolor{TableGray} \multicolumn{1}{|p{\SummaryTableFirstWidth cm}|}{\multirow{2}{*}{}} & \multicolumn{1}{|c|}{96.55\%} & \multicolumn{1}{|c|}{96.06\%} & \multicolumn{1}{|c|}{96.06\%} & \multicolumn{1}{|c|}{\textbf{54.29\%}}\\
    \rowcolor{TableGray} \multicolumn{1}{|p{\SummaryTableFirstWidth cm}|}{\multirow{-2}{*}{Success rate}} & \multicolumn{1}{|c|}{\(\sigma\)=0.18} & \multicolumn{1}{|c|}{\(\sigma\)=0.19, \(\chi^2\)=0.333, p=.564} & \multicolumn{1}{|c|}{\(\sigma\)=0.19, \(\chi^2\)=0.333, p=.564} & \multicolumn{1}{|c|}{\(\sigma\)=0.50, \(\chi^2\)=29.12, p<.001}\\
  \multicolumn{1}{|p{\SummaryTableFirstWidth cm}|}{\multirow{2}{*}{}} & \multicolumn{1}{|c|}{0.11m} & \multicolumn{1}{|c|}{0.13m} & \multicolumn{1}{|c|}{0.12m} & \multicolumn{1}{|c|}{\textbf{0.26m}}\\
    \multicolumn{1}{|p{\SummaryTableFirstWidth cm}|}{\multirow{-2}{*}{Position error}} & \multicolumn{1}{|c|}{\(\sigma\)=0.06} & \multicolumn{1}{|c|}{\(\sigma\)=0.17, t=1.162, p=.247} & \multicolumn{1}{|c|}{\(\sigma\)=0.16, t=0.855, p=.394} & \multicolumn{1}{|c|}{\(\sigma\)=0.19, t=7.153, p<.001}\\
  \rowcolor{TableGray} \multicolumn{1}{|p{\SummaryTableFirstWidth cm}|}{\multirow{2}{*}{}} & \multicolumn{1}{|c|}{5.97 deg} & \multicolumn{1}{|c|}{7.11 deg} & \multicolumn{1}{|c|}{7.87 deg} & \multicolumn{1}{|c|}{\textbf{17.56 deg}} \\
        \rowcolor{TableGray} \multicolumn{1}{|p{\SummaryTableFirstWidth cm}|}{\multirow{-2}{*}{Yaw error}} & \multicolumn{1}{|c|}{\(\sigma\)=8.78} & \multicolumn{1}{|c|}{\(\sigma\)=13.50, t=0.514, p=.608} & \multicolumn{1}{|c|}{\(\sigma\)=16.80, t=1.014, p=.312} & \multicolumn{1}{|c|}{\(\sigma\)=24.04, t=3.900, p<.001}\\
  \multicolumn{1}{|p{\SummaryTableFirstWidth cm}|}{\multirow{2}{*}{}} & \multicolumn{1}{|c|}{11.11s/m} & \multicolumn{1}{|c|}{\textbf{8.94s/m}} & \multicolumn{1}{|c|}{12.14s/m} & \multicolumn{1}{|c|}{11.38s/m} \\
    \multicolumn{1}{|p{\SummaryTableFirstWidth cm}|}{\multirow{-2}{*}{*Time / init. dist.}} & \multicolumn{1}{|c|}{\(\sigma\)=7.97} & \multicolumn{1}{|c|}{\(\sigma\)=5.44, t=3.618, p<.001} & \multicolumn{1}{|c|}{\(\sigma\)=9.02, t=1.159, p=.248} & \multicolumn{1}{|c|}{\(\sigma\)=6.50, t=0.817, p=.417}\\
  \rowcolor{TableGray} \multicolumn{1}{|p{\SummaryTableFirstWidth cm}|}{\multirow{2}{*}{}} & \multicolumn{1}{|c|}{1.64} & \multicolumn{1}{|c|}{\textbf{1.34}} & \multicolumn{1}{|c|}{1.96} & \multicolumn{1}{|c|}{1.38} \\
    \rowcolor{TableGray} \multicolumn{1}{|p{\SummaryTableFirstWidth cm}|}{\multirow{-2}{*}{*Dist. traveled / init. dist.}} & \multicolumn{1}{|c|}{\(\sigma\)=1.08} & \multicolumn{1}{|c|}{\(\sigma\)=0.72, t=3.400, p<.001} & \multicolumn{1}{|c|}{\(\sigma\)=1.39, t=2.551, p=.012} & \multicolumn{1}{|c|}{\(\sigma\)=0.96, t=0.910, p=.367}\\
  \hline
  
\end{tabular}
\end{center}
\begin{flushleft}
\footnotesize
Bolded values represent a statically significant difference compared to the Assisted condition at significance level \(\alpha = 0.01\). \\
A dependant t-test was used for all metrics aside from the success rate which used a McNemar's test. \\
*A learning bias was observed for the given metrics where participants were observed to require more time and generate less efficient trajectories within their first condition. The order for the Assisted, Assisted + Lights and Assisted + HoloLens was randomly assigned whilst the unassisted condition was always performed last. \\
For the Time / init. dist. and Dist. traveled / init. dist. metrics, the initial distance is calculated as the Euclidean distance from the starting point of the given task (the finishing point of the prior task) to that of the goal platform for the current task. Both time and distance traveled metrics account for the time taken and the distance traveled for the current task and not the entire mission. \\
Target approach is defined as the period in which the drone is traveling towards the intended goal until it first becomes within a threshold XY distance of 0.5m.
\end{flushleft}
\end{table}

\subsubsection{Task Performance Analysis}
The success rate for both landing and inspection tasks remained consistent through all three assisted conditions, averaging a 96\% task success rate, compared to the landing and inspection success rates in the unassisted condition of [16.67\% \& 56.06\%]. Compared to our previous work \cite{Kal2} where the UAV’s yaw remained locked and did not require participants to align the orientation of the UAV to that of the platform, the average unassisted success rate for landing was 51.43\%. The introduction of orientation control greatly decreased unassisted landing success to 32.41\% of the previous success rate, with 50\% of unassisted participants unable to achieve a single successful landing.  These results provide strong evidence that the assistant, trained only in simulation, is able to effectively infer pilot intent and assist diverse pilots to successfully complete challenging tasks.

\subsubsection{Task Performance Statistical Analysis}
Two tailed dependent t-tests under a 99\% confidence interval were performed to test for statistical significance amongst the three assisted conditions for metrics outlined in Table~\ref{UserStudySummaryTable}, excluding the success rate where a McNemar’s test was used under a 99\% confidence interval. There was insufficient evidence to suggest that either assisted condition had a statistically significant greater performance for the success, position and yaw errors over the other assisted conditions for both landing and inspection tasks. This is as expected due to the assistant being identical in each condition where the assistant’s aim is to ensure task success. However for the time and distance metrics, a statistically significant difference was found between the Assisted and Assisted + Lights conditions for the inspection task, indicating that the presence of red / green light feedback cues led to an overall reduction in the average time taken by 19.41\% and distance travelled by 17.80\% for the inspection task. The remaining time and distance metrics were found to be statistically insignificant.  
\subsection{Regression Analysis}
\subsubsection{Learning Effect and Yaw Control}
\begin{figure}[b]
\centering
\includegraphics[width=1.0\columnwidth]{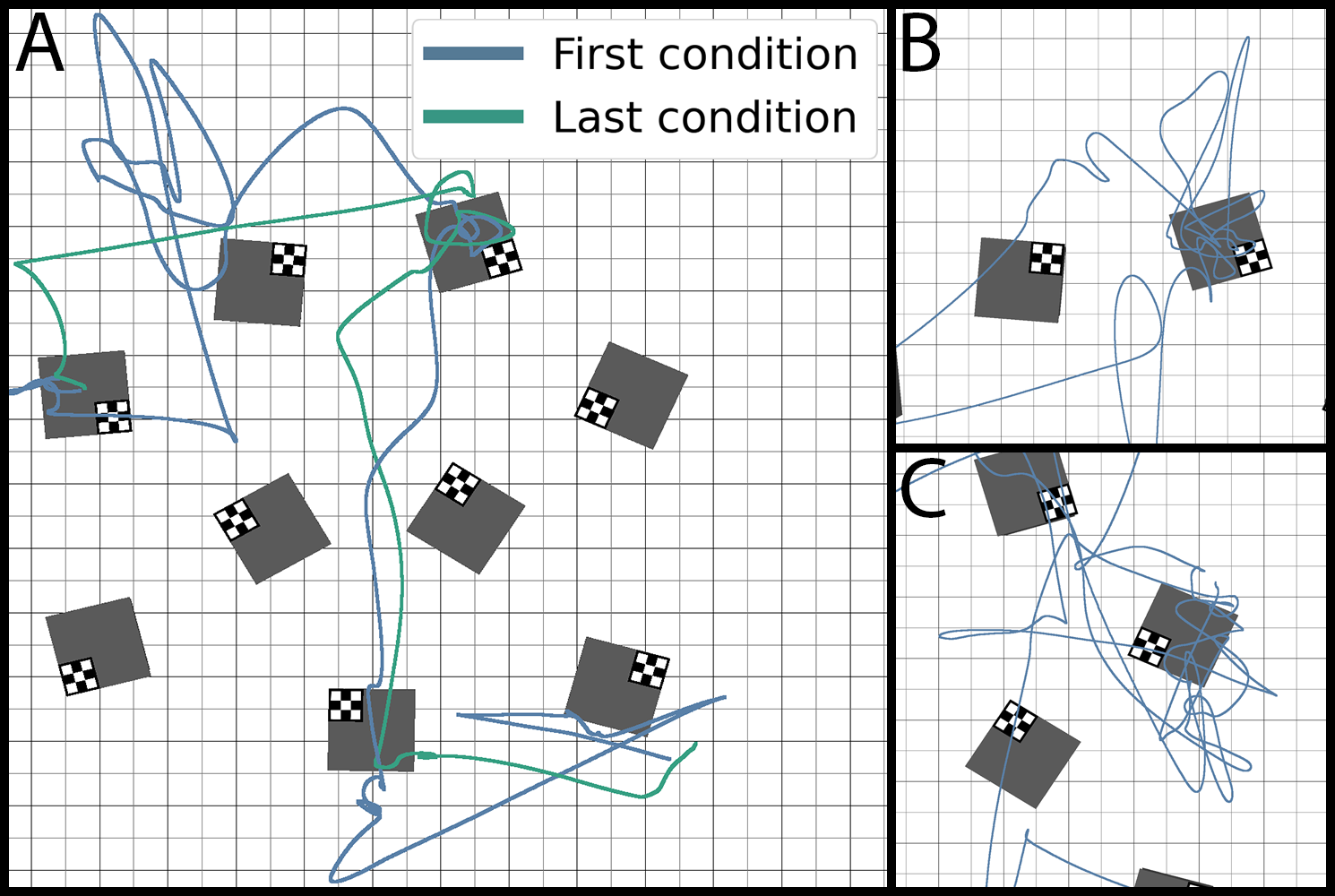}
\caption{Example trajectories of participants during the user study. 
Image A compares trajectories generated in the first and last condition where the participant initially generates an inefficient trajectory needing multiple adjustments while in the last condition they are able to fly directly to each of the goal platforms.
Images B and C demonstrates participants getting confused with the orientation of the drone, causing them to consistently fly in unintended directions. 
}
\label{ExampleTrajectory}
\end{figure}

Compared to our prior work \cite{Kal2}, the introduction of yaw control caused a strong learning effect as participants progressed further into the study, which was observed through progressively efficient trajectories and task completion times. 
An example trajectory is shown in Fig.~\ref{ExampleTrajectory} where in the first condition, participants often mistake the current heading of the UAV causing them to fly in an unintended direction. When attempting the same mission for the third condition they become accustomed to the UAV control scheme and generate a more efficient trajectory. 

The observed learning effect is empirically supported through regression analysis where a multi-variable regression model is formulated using the model: \(Y = A + AL + AH + CO\). The model’s output Y regresses six dependant variables: the time taken / initial task distance for the approach, inspection and landing task, and the distance traveled / initial task distance for the approach, inspection and landing task. The model uses four input independent variables where \(A\), \(AL\) and \(AH\) are binary variables indicating what condition the mission was flown in: Assisted, Assisted + Lights and Assisted + HoloLens respectively. The independent variables \(A\), \(AL\) \& \(AH\) aim to control for the impact each study condition may have on the time and distance metrics. The final independent variable \(CO\) controls for the order in which the condition was performed and can take a value of 0, 1 or 2 depending when a given condition was performed during the study. The aim of the model is to determine whether a learning effect can be observed by analyzing the effect \(CO\) has on the time and distance metrics.

Significance testing was performed on the coefficient affiliated with the \(CO\) independent variable to confirm if the condition order has an impact on the aforementioned time and distance traveled metrics. The \(CO\) variable was considered statistically significant under a two-tailed t-test of significance level \(\alpha\) = 0.01 for all six metrics. For all six metrics it was predicted that the further into the study a participant is, on average their expected task completion time and expected task travel distance is shorter. The full regression results can be seen in Table~\ref{RegressionSummaryLearning} in Appendix \ref{LearningEffectAppendix}. 

\subsubsection{Pilot Proficiency: Performance Influence}
Regression analysis is performed to assess whether a participant’s prior experience affects their assisted performance, preference to each of the assisted conditions and how much assistance the assistant provides. 
The model was formulated using: \(Y = \beta + P + k\), where independent variable \(\beta \) represents a participant’s UAV piloting proficiency and is calculated using the linear regression model outlined in \cite{Kal2} which uses participants’ responses from the demographics survey seen in Table~\ref{DemographicsTable}. Independent variable \(P\) denotes a prior participant of our previous user study \cite{Kal2}, while $k$ represents the intercept constant. 

The model was fitted to the performance metrics in Table \ref{UserStudySummaryTable} as dependant variables. Independent variable \(\beta \) was found to have a statistically significant influence on the Time / init. dist and Dist. travelled / init. Dist metrics, but not on the success, position or yaw error metrics of the task using a two-tailed t-test of significance level \(\alpha\) = 0.01. It was found that an increase in pilot’s proficiency was on average observed to reduce the time required and distance travelled to approach, land and complete an inspection task. 
There was insufficient evidence to suggest that a participant's prior participation (\(P\)) affected the success, position, yaw, time and distance traveled metrics of the task.

These results are partially congruent with our prior work \cite{Kal2} showing task success and position error to be invariant of piloting proficiency whilst assisted. However the inverse result is found for the time and distance travelled metrics, where a pilot’s proficiency in \cite{Kal2} was found to have no statistically significant effect on how fast or how efficient a trajectory a participant generated is whilst assisted. However in the current study, \(\beta \) was found to be statistically significant in estimating a pilot’s average time and distance travelled to complete a task. This is primarily due to the introduction of yaw control where novice pilots tended to get confused with the orientation of the UAV in the approach stage before the assistant could receive any cues about their intent. This ambiguity of inferring the pilot’s intent is further explored in Section \ref{TaskAssistanceSection}. 

\subsubsection{Pilot Proficiency: Condition Preference}
Regression analysis was then performed using the previous \(Y = \beta + P + k\) regression model to estimate whether a pilot’s expertise influenced their preferred assistance condition. Participants’ response to question 4 as shown in Table~\ref{WordedResponse} was used as a dependent variable to estimate pilot preferences. There was insufficient evidence to suggest that \(\beta \) or \(P\) impacted pilots' preferences to either of the experimental conditions under a 99\% confidence interval.

\begin{table}[t]
\begin{center}
\captionof{table}{Pilot proficiency regression model}
\newcommand\RegressionTableOneFirstWidth{4.5}
\newcommand\RegressionTableOneSecondWidth{1.7}
\setlength{\tabcolsep}{0.2em}
\label{RegressionSummaryProficiency}
\definecolor{TableGray}{gray}{0.9}
\footnotesize
\begin{tabular}{|p{\RegressionTableOneFirstWidth cm}|c|c|}
  
  \hline
  \multicolumn{1}{|p{\RegressionTableOneFirstWidth cm}|}{} & \multicolumn{1}{|p{\RegressionTableOneSecondWidth cm}|}{\centering \(\beta\)} & \multicolumn{1}{|p{\RegressionTableOneSecondWidth cm}|}{\centering \(P\)} \\
  \hline
  \multicolumn{3}{|c|}{Target approach} \\ 
  \hline
  \multicolumn{1}{|p{\RegressionTableOneFirstWidth cm}|}{\multirow{1}{*}{Time / init. dist.}}           & \multicolumn{1}{|c|}{\textbf{-2.07}} & \multicolumn{1}{|c|}{-0.42} \\
  \rowcolor{TableGray}
  \multicolumn{1}{|p{\RegressionTableOneFirstWidth cm}|}{\multirow{1}{*}{Dist. traveled / init. dist.}} & \multicolumn{1}{|c|}{\textbf{-0.47}} & \multicolumn{1}{|c|}{-0.11} \\
  \hline
  \multicolumn{3}{|c|}{Landing task} \\ 
  \hline
  \rowcolor{TableGray}
  \multicolumn{1}{|p{\RegressionTableOneFirstWidth cm}|}{\multirow{1}{*}{Success rate}}                 & \multicolumn{1}{|c|}{ 0.08}           & \multicolumn{1}{|c|}{ 0.03}  \\
  \multicolumn{1}{|p{\RegressionTableOneFirstWidth cm}|}{\multirow{1}{*}{Position} error}                & \multicolumn{1}{|c|}{-0.03}           & \multicolumn{1}{|c|}{-0.02}  \\
  \rowcolor{TableGray}
  \multicolumn{1}{|p{\RegressionTableOneFirstWidth cm}|}{\multirow{1}{*}{Yaw error}}                    & \multicolumn{1}{|c|}{-0.03}           & \multicolumn{1}{|c|}{-0.02}  \\
  \multicolumn{1}{|p{\RegressionTableOneFirstWidth cm}|}{\multirow{1}{*}{Time / init. dist.}}           & \multicolumn{1}{|c|}{\textbf{-9.68}}  & \multicolumn{1}{|c|}{-0.96}  \\
  \rowcolor{TableGray}
  \multicolumn{1}{|p{\RegressionTableOneFirstWidth cm}|}{\multirow{1}{*}{Dist. traveled / init. dist.}} & \multicolumn{1}{|c|}{\textbf{-0.82}}  & \multicolumn{1}{|c|}{-0.25}  \\
  \hline
  \multicolumn{3}{|c|}{Inspection task} \\ 
  \hline
  \rowcolor{TableGray}
  \multicolumn{1}{|p{\RegressionTableOneFirstWidth cm}|}{\multirow{1}{*}{Success rate}}                 & \multicolumn{1}{|c|}{ 0.03}          & \multicolumn{1}{|c|}{ 0.04} \\
  \multicolumn{1}{|p{\RegressionTableOneFirstWidth cm}|}{\multirow{1}{*}{Position} error}             & \multicolumn{1}{|c|}{-0.01}          & \multicolumn{1}{|c|}{-0.01} \\
  \rowcolor{TableGray}
  \multicolumn{1}{|p{\RegressionTableOneFirstWidth cm}|}{\multirow{1}{*}{Yaw error}}                    & \multicolumn{1}{|c|}{-0.06}          & \multicolumn{1}{|c|}{-0.04} \\
  \multicolumn{1}{|p{\RegressionTableOneFirstWidth cm}|}{\multirow{1}{*}{Time / init. dist.}}           & \multicolumn{1}{|c|}{\textbf{-8.26}} & \multicolumn{1}{|c|}{-0.86} \\
  \rowcolor{TableGray}
  \multicolumn{1}{|p{\RegressionTableOneFirstWidth cm}|}{\multirow{1}{*}{Dist. traveled / init. dist.}} & \multicolumn{1}{|c|}{\textbf{-0.90}} & \multicolumn{1}{|c|}{-0.19} \\
  \hline
  \multicolumn{3}{|c|}{Condition preference} \\ 
  \hline
  \rowcolor{TableGray}
  \multicolumn{1}{|p{\RegressionTableOneFirstWidth cm}|}{\multirow{1}{*}{Assisted ranking}}                 & \multicolumn{1}{|c|}{-0.42}      & \multicolumn{1}{|c|}{-0.01}   \\
  \multicolumn{1}{|p{\RegressionTableOneFirstWidth cm}|}{\multirow{1}{*}{Assisted + Lights ranking}}        & \multicolumn{1}{|c|}{ 0.69}      & \multicolumn{1}{|c|}{ 0.26}   \\
  \rowcolor{TableGray}
  \multicolumn{1}{|p{\RegressionTableOneFirstWidth cm}|}{\multirow{1}{*}{Assisted + HoloLens ranking}}      & \multicolumn{1}{|c|}{-0.27}      & \multicolumn{1}{|c|}{-0.25}   \\
  \hline
  \multicolumn{3}{|c|}{Assistance received} \\ 
  \hline
   \rowcolor{TableGray}
  \multicolumn{1}{|p{\RegressionTableOneFirstWidth cm}|}{\multirow{1}{*}{XYZ assistance}}                  & \multicolumn{1}{|c|}{\textbf{-271.49}}  & \multicolumn{1}{|c|}{\textbf{-140.66}}  \\
  \multicolumn{1}{|p{\RegressionTableOneFirstWidth cm}|}{\multirow{1}{*}{Yaw assistance}}                  & \multicolumn{1}{|c|}{\textbf{-549.95}}  & \multicolumn{1}{|c|}{-59.99}            \\
  \hline

\end{tabular}
\begin{flushleft}
\footnotesize
Regression coefficients for the model \(Y = \beta + P + k\). \\
Regression constant \(k\) omitted for clarity. \\
Bolded values represent statistically significant values under a two-tailed t-test of significance \(\alpha = 0.01\).    
\end{flushleft}
\end{center}
\end{table}

\subsubsection{Pilot Proficiency: Assistance Received}
The regression model \(Y = \beta + P + k\) was finally used to determine if pilot proficiency influences the total assistance provided by the assistant, where the assistance received along the XYZ and yaw axis was used as dependent variables. It was found that both \(\beta\) and \(P\) were statistically significant in estimating the amount of assistance provided for the linear XYZ assistance, where an increase in pilot proficiency or if the participant had participated in the prior study, decreased the overall assistance required to complete the task.

For the assistance received along the yaw axis, only pilot proficiency (\(\beta \)) was found to be statistically significant under a 99\% confidence interval, where an increase in pilot proficiency reduced the amount of assistance required along the yaw axis. This is expected as prior participants did not experience yaw control in the previous study. These results suggest that the assistant appropriately adjusted the level of assistance based on the pilot's capability. The full regression results can be seen in Table~\ref{RegressionSummaryProficiency}.

\subsection{Task Assistance}\label{TaskAssistanceSection}
\begin{figure}
\centering
\includegraphics[width=1.0\columnwidth]{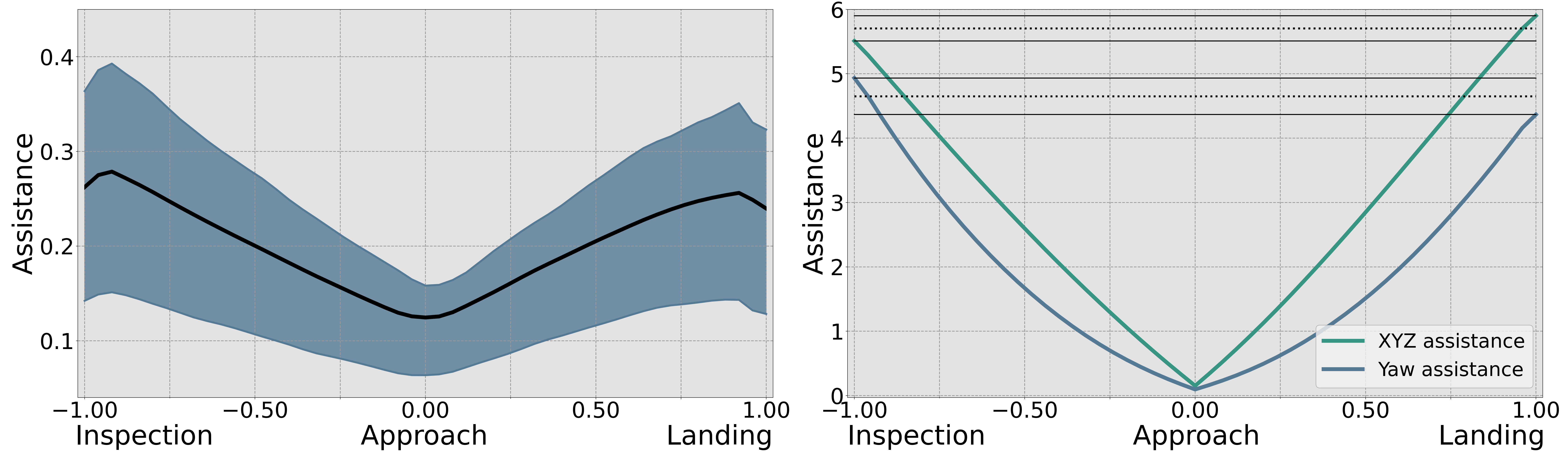}
\caption{Assistance received for each task and transitions between tasks. (Left-figure) Average assistance received (black) and interquartile range of assistance received (blue region). (Right-figure) Cumulative average assistance received for inspection and landing tasks split along the XYZ axis (green) and yaw axis (blue). Assistance (y-axis) is defined as the magnitude difference between the pilot’s action and the assistant’s action. For average assistance, both the XYZ and yaw assistance are averaged. The transition from the approach to the landing or inspection task (x-axis) is formulated by normalizing the time length of the task between 0 and 1 and bucketing the assistance values into 51 buckets in which the average is taken for each bucket.}
\label{TaskAssistance}
\end{figure}

The degree of assistance provided by the assistant, defined as the magnitude of the difference between the pilot's action and the assistant's action is dependent on the confidence in the estimated intent of the pilot. During the approach, the ability of the assistant to estimate the desired goal is low due to the presence of multiple potential goals with little indication from the pilot of targeted actions focusing on a single goal. As the task transitions from approach to the goal objective (inspection or landing), the ability to estimate the pilot’s intent increases, hence an increase in the assistance provided is observed as seen in the left plot in Fig.~\ref{TaskAssistance}.

The cumulative average assistance for both landing and inspection tasks are equal, where the landing task is observed to have a greater bias towards assistance along the XYZ direction due to the precise success requirements in the XY position of the UAV for landing compared to inspection. While the inspection task was observed to have a greater need towards assistance along the yaw direction compared to the landing task as seen in the right plot in Fig.~\ref{TaskAssistance}. The higher observed yaw assistance is due to precise alignment with the target platform maximises platform image coverage whilst minimising risk of task failure from corners being omitted when an image is taken. 

\subsection{Participant Perception Results}\label{UserStudyResultsQualitative}
\begin{figure}[b]
\centering
\includegraphics[width=1.0\columnwidth]{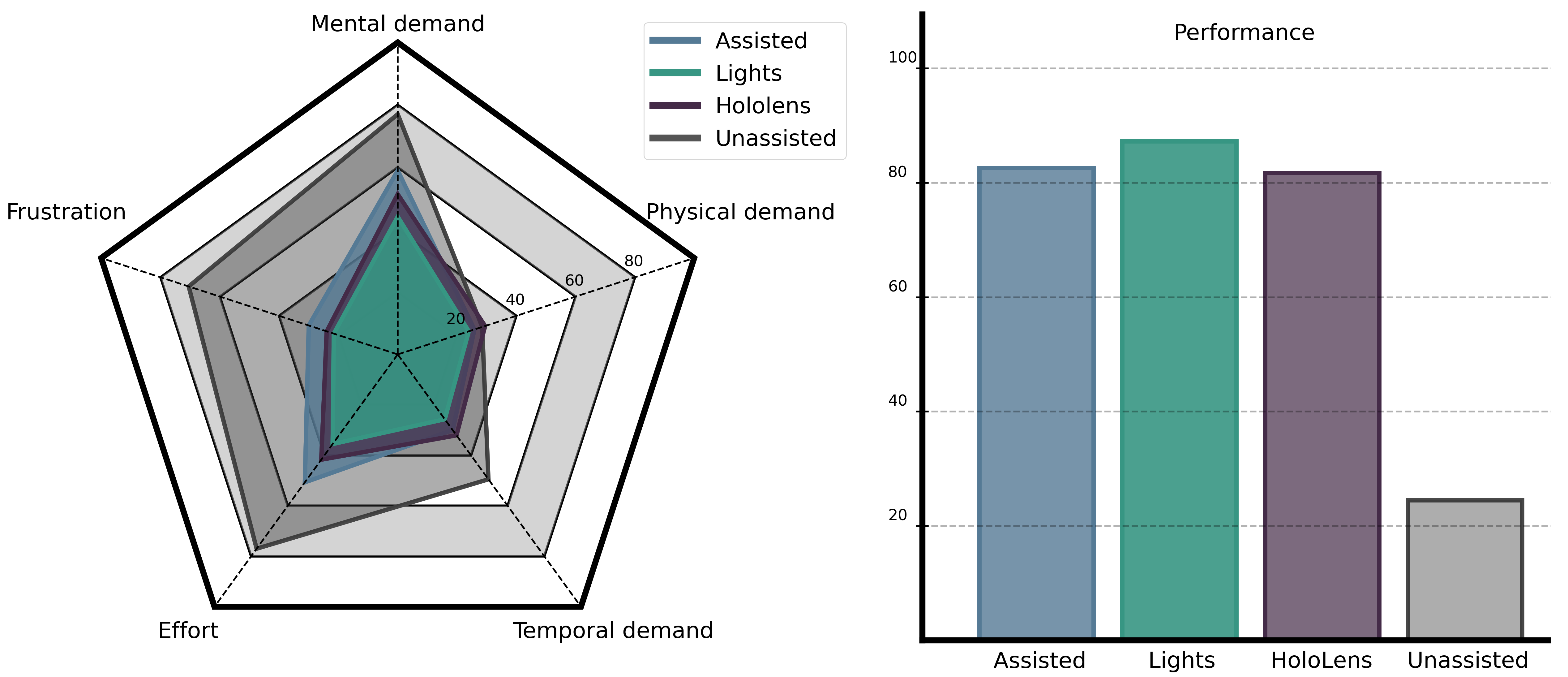}
\caption{Results of NASA TLX survey.}
\label{TLXResults}
\end{figure}

Summary results of participants’ perception to each of the conditions can be seen in Fig.~\ref{TLXResults} and Table~\ref{ConditionPerceptionTable}.

\subsubsection{Task Load Perception Statistical Analysis}
The TLX survey responses were analysed with a Welch’s t-test under a 95\% confidence interval to determine if a statistical difference exists between any of the study conditions for each of the workload metrics. A statistically significant difference was found between the assisted condition and the unassisted condition for each of the metrics aside from physical demand. A statistically significant difference was also found between the Assisted condition and the Assisted + Lights condition for the mental demand and effort workload metric, indicating that the inclusion of light feedback reduced pilots' mental workload and effort required to complete the task. For all other metrics compared to the assisted condition, there was insufficient evidence to suggest that a statistically significant difference exists. 

\subsubsection{Condition Perception Analysis}
From the condition perception survey as shown in Table~\ref{ConditionPerceptionTable}, it was found that compared to the assisted condition, introducing the ability for the assistant to provide feedback via flashing red and green lights enhanced pilot’s confidence in the landing and inspection task. Furthermore allowing the assistant to provide feedback created a greater understanding between the pilot and the assistant, where pilots better understood what the intent of the assistant was and what the assistant expected of them. 

\begin{figure}[!t]
 \captionof{table}{Condition perception survey}
 \label{ConditionPerceptionTable}
 \begin{center}
 \begin{tikzpicture}
\matrix [ConditionPerceptionTable,text width=3.0em] (M)
{
&A&A+L&A+H\\
\textbf1: I felt confident in completing the landing task&&&\\
\textbf2: I felt confident in completing the inspection task&&&\\
\textbf3: I understood what the assistant wanted to do&&&\\
\textbf4: I understood what the assistant wanted me to do&&&\\
\textbf5: The assistant didn’t do what I wanted it to do&&&\\
\textbf6: The assistant and I worked well as a team&&&\\
\textbf7: I felt that I received a lot of help from the assistant&&&\\
};

\definecolor{OddColor}{rgb}{0.3333333333333333, 0.47843137254901963, 0.5843137254901961}
\definecolor{EvenColor}{rgb}{0.21568627450980393, 0.5882352941176471, 0.5137254901960784}

\DrawCell{M-1-1.north west}{M-1-1.north east}{M-1-1.south east}{M-1-1.south west}
\DrawCell{M-1-2.north west}{M-1-2.north east}{M-1-2.south east}{M-1-2.south west}
\DrawCell{M-1-3.north west}{M-1-3.north east}{M-1-3.south east}{M-1-3.south west}
\DrawCell{M-1-4.north west}{M-1-4.north east}{M-1-4.south east}{M-1-4.south west}

\DrawCell{M-2-1.north west}{M-2-1.north east}{M-2-1.south east}{M-2-1.south west}
\DrawCell{M-2-2.north west}{M-2-2.north east}{M-2-2.south east}{M-2-2.south west}
\DrawCell{M-2-3.north west}{M-2-3.north east}{M-2-3.south east}{M-2-3.south west}
\DrawCell{M-2-4.north west}{M-2-4.north east}{M-2-4.south east}{M-2-4.south west}

\DrawCell{M-3-1.north west}{M-3-1.north east}{M-3-1.south east}{M-3-1.south west}
\DrawCell{M-3-2.north west}{M-3-2.north east}{M-3-2.south east}{M-3-2.south west}
\DrawCell{M-3-3.north west}{M-3-3.north east}{M-3-3.south east}{M-3-3.south west}
\DrawCell{M-3-4.north west}{M-3-4.north east}{M-3-4.south east}{M-3-4.south west}

\DrawCell{M-4-1.north west}{M-4-1.north east}{M-4-1.south east}{M-4-1.south west}
\DrawCell{M-4-2.north west}{M-4-2.north east}{M-4-2.south east}{M-4-2.south west}
\DrawCell{M-4-3.north west}{M-4-3.north east}{M-4-3.south east}{M-4-3.south west}
\DrawCell{M-4-4.north west}{M-4-4.north east}{M-4-4.south east}{M-4-4.south west}

\DrawCell{M-5-1.north west}{M-5-1.north east}{M-5-1.south east}{M-5-1.south west}
\DrawCell{M-5-2.north west}{M-5-2.north east}{M-5-2.south east}{M-5-2.south west}
\DrawCell{M-5-3.north west}{M-5-3.north east}{M-5-3.south east}{M-5-3.south west}
\DrawCell{M-5-4.north west}{M-5-4.north east}{M-5-4.south east}{M-5-4.south west}

\DrawCell{M-6-1.north west}{M-6-1.north east}{M-6-1.south east}{M-6-1.south west}
\DrawCell{M-6-2.north west}{M-6-2.north east}{M-6-2.south east}{M-6-2.south west}
\DrawCell{M-6-3.north west}{M-6-3.north east}{M-6-3.south east}{M-6-3.south west}
\DrawCell{M-6-4.north west}{M-6-4.north east}{M-6-4.south east}{M-6-4.south west}

\DrawCell{M-7-1.north west}{M-7-1.north east}{M-7-1.south east}{M-7-1.south west}
\DrawCell{M-7-2.north west}{M-7-2.north east}{M-7-2.south east}{M-7-2.south west}
\DrawCell{M-7-3.north west}{M-7-3.north east}{M-7-3.south east}{M-7-3.south west}
\DrawCell{M-7-4.north west}{M-7-4.north east}{M-7-4.south east}{M-7-4.south west}

\DrawCell{M-8-1.north west}{M-8-1.north east}{M-8-1.south east}{M-8-1.south west}
\DrawCell{M-8-2.north west}{M-8-2.north east}{M-8-2.south east}{M-8-2.south west}
\DrawCell{M-8-3.north west}{M-8-3.north east}{M-8-3.south east}{M-8-3.south west}
\DrawCell{M-8-4.north west}{M-8-4.north east}{M-8-4.south east}{M-8-4.south west}

\node[anchor = north west, align=center, font=\footnotesize] at (M-2-2.north west) {\\ \hspace{0.17em} \textbf{ 3.14}};
\node[anchor = north west, align=center, font=\footnotesize] at (M-2-3.north west) {\\ \hspace{0.17em} \textbf{ 3.55}};
\node[anchor = north west, align=center, font=\footnotesize] at (M-2-4.north west) {\\ \hspace{0.17em} \textbf{ 3.28}};

\node[anchor = north west, align=center, font=\footnotesize] at (M-3-2.north west) {\\ \hspace{0.17em} \textbf{ 3.07}};
\node[anchor = north west, align=center, font=\footnotesize] at (M-3-3.north west) {\\ \hspace{0.17em} \textbf{ 3.52}};
\node[anchor = north west, align=center, font=\footnotesize] at (M-3-4.north west) {\\ \hspace{0.17em} \textbf{ 3.07}};

\node[anchor = north west, align=center, font=\footnotesize] at (M-4-2.north west) {\\ \hspace{0.17em} \textbf{ 3.10}};
\node[anchor = north west, align=center, font=\footnotesize] at (M-4-3.north west) {\\ \hspace{0.17em} \textbf{ 3.62}};
\node[anchor = north west, align=center, font=\footnotesize] at (M-4-4.north west) {\\ \hspace{0.17em} \textbf{ 3.14}};

\node[anchor = north west, align=center, font=\footnotesize] at (M-5-2.north west) {\\ \hspace{0.17em} \textbf{ 2.45}};
\node[anchor = north west, align=center, font=\footnotesize] at (M-5-3.north west) {\\ \hspace{0.17em} \textbf{ 3.38}};
\node[anchor = north west, align=center, font=\footnotesize] at (M-5-4.north west) {\\ \hspace{0.17em} \textbf{ 2.83}};

\node[anchor = north west, align=center, font=\footnotesize] at (M-6-2.north west) {\\ \hspace{0.17em} \textbf{ 1.03}};
\node[anchor = north west, align=center, font=\footnotesize] at (M-6-3.north west) {\\ \hspace{0.17em} \textbf{ 0.97}};
\node[anchor = north west, align=center, font=\footnotesize] at (M-6-4.north west) {\\ \hspace{0.17em} \textbf{ 0.93}};

\node[anchor = north west, align=center, font=\footnotesize] at (M-7-2.north west) {\\ \hspace{0.17em} \textbf{ 2.97}};
\node[anchor = north west, align=center, font=\footnotesize] at (M-7-3.north west) {\\ \hspace{0.17em} \textbf{ 3.41}};
\node[anchor = north west, align=center, font=\footnotesize] at (M-7-4.north west) {\\ \hspace{0.17em} \textbf{ 3.07}};

\node[anchor = north west, align=center, font=\footnotesize] at (M-8-2.north west) {\\ \hspace{0.17em} \textbf{ 3.00}};
\node[anchor = north west, align=center, font=\footnotesize] at (M-8-3.north west) {\\ \hspace{0.17em} \textbf{ 3.31}};
\node[anchor = north west, align=center, font=\footnotesize] at (M-8-4.north west) {\\ \hspace{0.17em} \textbf{ 3.07}};

\draw[opacity=0.6,fill=OddColor]($ (M-2-2.south west) + (0em, 0em) $) -- ($ (M-2-2.south west) + (0.0em, 0.0em) $) -- ($ (M-2-2.south west) + (0.7400000000000001em, 0.0em) $) -- ($ (M-2-2.south west) + (0.7400000000000001em, 0.09568965517241379em) $) -- ($ (M-2-2.south west) + (1.4800000000000002em, 0.09568965517241379em) $) -- ($ (M-2-2.south west) + (1.4800000000000002em, 0.14353448275862069em) $) -- ($ (M-2-2.south west) + (2.2200000000000006em, 0.14353448275862069em) $) -- ($ (M-2-2.south west) + (2.2200000000000006em, 0.6219827586206896em) $) -- ($ (M-2-2.south west) + (2.9600000000000004em, 0.6219827586206896em) $) -- ($ (M-2-2.south west) + (2.9600000000000004em, 0.5262931034482758em) $) -- ($ (M-2-2.south west) + (3.7em, 0.5262931034482758em) $) -- ($ (M-2-2.south west) + (3.7em, 0em) $) -- cycle;
\draw[opacity=0.6,fill=OddColor]($ (M-2-3.south west) + (0em, 0em) $) -- ($ (M-2-3.south west) + (0.0em, 0.0em) $) -- ($ (M-2-3.south west) + (0.7400000000000001em, 0.0em) $) -- ($ (M-2-3.south west) + (0.7400000000000001em, 0.0em) $) -- ($ (M-2-3.south west) + (1.4800000000000002em, 0.0em) $) -- ($ (M-2-3.south west) + (1.4800000000000002em, 0.047844827586206895em) $) -- ($ (M-2-3.south west) + (2.2200000000000006em, 0.047844827586206895em) $) -- ($ (M-2-3.south west) + (2.2200000000000006em, 0.5262931034482758em) $) -- ($ (M-2-3.south west) + (2.9600000000000004em, 0.5262931034482758em) $) -- ($ (M-2-3.south west) + (2.9600000000000004em, 0.8133620689655172em) $) -- ($ (M-2-3.south west) + (3.7em, 0.8133620689655172em) $) -- ($ (M-2-3.south west) + (3.7em, 0em) $) -- cycle;
\draw[opacity=0.6,fill=OddColor]($ (M-2-4.south west) + (0em, 0em) $) -- ($ (M-2-4.south west) + (0.0em, 0.0em) $) -- ($ (M-2-4.south west) + (0.7400000000000001em, 0.0em) $) -- ($ (M-2-4.south west) + (0.7400000000000001em, 0.047844827586206895em) $) -- ($ (M-2-4.south west) + (1.4800000000000002em, 0.047844827586206895em) $) -- ($ (M-2-4.south west) + (1.4800000000000002em, 0.14353448275862069em) $) -- ($ (M-2-4.south west) + (2.2200000000000006em, 0.14353448275862069em) $) -- ($ (M-2-4.south west) + (2.2200000000000006em, 0.5741379310344827em) $) -- ($ (M-2-4.south west) + (2.9600000000000004em, 0.5741379310344827em) $) -- ($ (M-2-4.south west) + (2.9600000000000004em, 0.6219827586206896em) $) -- ($ (M-2-4.south west) + (3.7em, 0.6219827586206896em) $) -- ($ (M-2-4.south west) + (3.7em, 0em) $) -- cycle;

\draw[opacity=0.6,fill=EvenColor]($ (M-3-2.south west) + (0em, 0em) $) -- ($ (M-3-2.south west) + (0.0em, 0.0em) $) -- ($ (M-3-2.south west) + (0.7400000000000001em, 0.0em) $) -- ($ (M-3-2.south west) + (0.7400000000000001em, 0.047844827586206895em) $) -- ($ (M-3-2.south west) + (1.4800000000000002em, 0.047844827586206895em) $) -- ($ (M-3-2.south west) + (1.4800000000000002em, 0.19137931034482758em) $) -- ($ (M-3-2.south west) + (2.2200000000000006em, 0.19137931034482758em) $) -- ($ (M-3-2.south west) + (2.2200000000000006em, 0.7655172413793103em) $) -- ($ (M-3-2.south west) + (2.9600000000000004em, 0.7655172413793103em) $) -- ($ (M-3-2.south west) + (2.9600000000000004em, 0.38275862068965516em) $) -- ($ (M-3-2.south west) + (3.7em, 0.38275862068965516em) $) -- ($ (M-3-2.south west) + (3.7em, 0em) $) -- cycle;
\draw[opacity=0.6,fill=EvenColor]($ (M-3-3.south west) + (0em, 0em) $) -- ($ (M-3-3.south west) + (0.0em, 0.0em) $) -- ($ (M-3-3.south west) + (0.7400000000000001em, 0.0em) $) -- ($ (M-3-3.south west) + (0.7400000000000001em, 0.0em) $) -- ($ (M-3-3.south west) + (1.4800000000000002em, 0.0em) $) -- ($ (M-3-3.south west) + (1.4800000000000002em, 0.09568965517241379em) $) -- ($ (M-3-3.south west) + (2.2200000000000006em, 0.09568965517241379em) $) -- ($ (M-3-3.south west) + (2.2200000000000006em, 0.478448275862069em) $) -- ($ (M-3-3.south west) + (2.9600000000000004em, 0.478448275862069em) $) -- ($ (M-3-3.south west) + (2.9600000000000004em, 0.8133620689655172em) $) -- ($ (M-3-3.south west) + (3.7em, 0.8133620689655172em) $) -- ($ (M-3-3.south west) + (3.7em, 0em) $) -- cycle;
\draw[opacity=0.6,fill=EvenColor]($ (M-3-4.south west) + (0em, 0em) $) -- ($ (M-3-4.south west) + (0.0em, 0.0em) $) -- ($ (M-3-4.south west) + (0.7400000000000001em, 0.0em) $) -- ($ (M-3-4.south west) + (0.7400000000000001em, 0.0em) $) -- ($ (M-3-4.south west) + (1.4800000000000002em, 0.0em) $) -- ($ (M-3-4.south west) + (1.4800000000000002em, 0.3349137931034483em) $) -- ($ (M-3-4.south west) + (2.2200000000000006em, 0.3349137931034483em) $) -- ($ (M-3-4.south west) + (2.2200000000000006em, 0.6219827586206896em) $) -- ($ (M-3-4.south west) + (2.9600000000000004em, 0.6219827586206896em) $) -- ($ (M-3-4.south west) + (2.9600000000000004em, 0.43060344827586217em) $) -- ($ (M-3-4.south west) + (3.7em, 0.43060344827586217em) $) -- ($ (M-3-4.south west) + (3.7em, 0em) $) -- cycle;

\draw[opacity=0.6,fill=OddColor]($ (M-4-2.south west) + (0em, 0em) $) -- ($ (M-4-2.south west) + (0.0em, 0.047844827586206895em) $) -- ($ (M-4-2.south west) + (0.7400000000000001em, 0.047844827586206895em) $) -- ($ (M-4-2.south west) + (0.7400000000000001em, 0.047844827586206895em) $) -- ($ (M-4-2.south west) + (1.4800000000000002em, 0.047844827586206895em) $) -- ($ (M-4-2.south west) + (1.4800000000000002em, 0.14353448275862069em) $) -- ($ (M-4-2.south west) + (2.2200000000000006em, 0.14353448275862069em) $) -- ($ (M-4-2.south west) + (2.2200000000000006em, 0.6219827586206896em) $) -- ($ (M-4-2.south west) + (2.9600000000000004em, 0.6219827586206896em) $) -- ($ (M-4-2.south west) + (2.9600000000000004em, 0.5262931034482758em) $) -- ($ (M-4-2.south west) + (3.7em, 0.5262931034482758em) $) -- ($ (M-4-2.south west) + (3.7em, 0em) $) -- cycle;
\draw[opacity=0.6,fill=OddColor]($ (M-4-3.south west) + (0em, 0em) $) -- ($ (M-4-3.south west) + (0.0em, 0.0em) $) -- ($ (M-4-3.south west) + (0.7400000000000001em, 0.0em) $) -- ($ (M-4-3.south west) + (0.7400000000000001em, 0.0em) $) -- ($ (M-4-3.south west) + (1.4800000000000002em, 0.0em) $) -- ($ (M-4-3.south west) + (1.4800000000000002em, 0.047844827586206895em) $) -- ($ (M-4-3.south west) + (2.2200000000000006em, 0.047844827586206895em) $) -- ($ (M-4-3.south west) + (2.2200000000000006em, 0.43060344827586217em) $) -- ($ (M-4-3.south west) + (2.9600000000000004em, 0.43060344827586217em) $) -- ($ (M-4-3.south west) + (2.9600000000000004em, 0.909051724137931em) $) -- ($ (M-4-3.south west) + (3.7em, 0.909051724137931em) $) -- ($ (M-4-3.south west) + (3.7em, 0em) $) -- cycle;
\draw[opacity=0.6,fill=OddColor]($ (M-4-4.south west) + (0em, 0em) $) -- ($ (M-4-4.south west) + (0.0em, 0.0em) $) -- ($ (M-4-4.south west) + (0.7400000000000001em, 0.0em) $) -- ($ (M-4-4.south west) + (0.7400000000000001em, 0.14353448275862069em) $) -- ($ (M-4-4.south west) + (1.4800000000000002em, 0.14353448275862069em) $) -- ($ (M-4-4.south west) + (1.4800000000000002em, 0.09568965517241379em) $) -- ($ (M-4-4.south west) + (2.2200000000000006em, 0.09568965517241379em) $) -- ($ (M-4-4.south west) + (2.2200000000000006em, 0.5741379310344827em) $) -- ($ (M-4-4.south west) + (2.9600000000000004em, 0.5741379310344827em) $) -- ($ (M-4-4.south west) + (2.9600000000000004em, 0.5741379310344827em) $) -- ($ (M-4-4.south west) + (3.7em, 0.5741379310344827em) $) -- ($ (M-4-4.south west) + (3.7em, 0em) $) -- cycle;

\draw[opacity=0.6,fill=EvenColor]($ (M-5-2.south west) + (0em, 0em) $) -- ($ (M-5-2.south west) + (0.0em, 0.09568965517241379em) $) -- ($ (M-5-2.south west) + (0.7400000000000001em, 0.09568965517241379em) $) -- ($ (M-5-2.south west) + (0.7400000000000001em, 0.19137931034482758em) $) -- ($ (M-5-2.south west) + (1.4800000000000002em, 0.19137931034482758em) $) -- ($ (M-5-2.south west) + (1.4800000000000002em, 0.38275862068965516em) $) -- ($ (M-5-2.south west) + (2.2200000000000006em, 0.38275862068965516em) $) -- ($ (M-5-2.south west) + (2.2200000000000006em, 0.43060344827586217em) $) -- ($ (M-5-2.south west) + (2.9600000000000004em, 0.43060344827586217em) $) -- ($ (M-5-2.south west) + (2.9600000000000004em, 0.28706896551724137em) $) -- ($ (M-5-2.south west) + (3.7em, 0.28706896551724137em) $) -- ($ (M-5-2.south west) + (3.7em, 0em) $) -- cycle;
\draw[opacity=0.6,fill=EvenColor]($ (M-5-3.south west) + (0em, 0em) $) -- ($ (M-5-3.south west) + (0.0em, 0.0em) $) -- ($ (M-5-3.south west) + (0.7400000000000001em, 0.0em) $) -- ($ (M-5-3.south west) + (0.7400000000000001em, 0.047844827586206895em) $) -- ($ (M-5-3.south west) + (1.4800000000000002em, 0.047844827586206895em) $) -- ($ (M-5-3.south west) + (1.4800000000000002em, 0.14353448275862069em) $) -- ($ (M-5-3.south west) + (2.2200000000000006em, 0.14353448275862069em) $) -- ($ (M-5-3.south west) + (2.2200000000000006em, 0.43060344827586217em) $) -- ($ (M-5-3.south west) + (2.9600000000000004em, 0.43060344827586217em) $) -- ($ (M-5-3.south west) + (2.9600000000000004em, 0.7655172413793103em) $) -- ($ (M-5-3.south west) + (3.7em, 0.7655172413793103em) $) -- ($ (M-5-3.south west) + (3.7em, 0em) $) -- cycle;
\draw[opacity=0.6,fill=EvenColor]($ (M-5-4.south west) + (0em, 0em) $) -- ($ (M-5-4.south west) + (0.0em, 0.047844827586206895em) $) -- ($ (M-5-4.south west) + (0.7400000000000001em, 0.047844827586206895em) $) -- ($ (M-5-4.south west) + (0.7400000000000001em, 0.19137931034482758em) $) -- ($ (M-5-4.south west) + (1.4800000000000002em, 0.19137931034482758em) $) -- ($ (M-5-4.south west) + (1.4800000000000002em, 0.14353448275862069em) $) -- ($ (M-5-4.south west) + (2.2200000000000006em, 0.14353448275862069em) $) -- ($ (M-5-4.south west) + (2.2200000000000006em, 0.5741379310344827em) $) -- ($ (M-5-4.south west) + (2.9600000000000004em, 0.5741379310344827em) $) -- ($ (M-5-4.south west) + (2.9600000000000004em, 0.43060344827586217em) $) -- ($ (M-5-4.south west) + (3.7em, 0.43060344827586217em) $) -- ($ (M-5-4.south west) + (3.7em, 0em) $) -- cycle;

\draw[opacity=0.6,fill=OddColor]($ (M-6-2.south west) + (0em, 0em) $) -- ($ (M-6-2.south west) + (0.0em, 0.5262931034482758em) $) -- ($ (M-6-2.south west) + (0.7400000000000001em, 0.5262931034482758em) $) -- ($ (M-6-2.south west) + (0.7400000000000001em, 0.5262931034482758em) $) -- ($ (M-6-2.south west) + (1.4800000000000002em, 0.5262931034482758em) $) -- ($ (M-6-2.south west) + (1.4800000000000002em, 0.14353448275862069em) $) -- ($ (M-6-2.south west) + (2.2200000000000006em, 0.14353448275862069em) $) -- ($ (M-6-2.south west) + (2.2200000000000006em, 0.14353448275862069em) $) -- ($ (M-6-2.south west) + (2.9600000000000004em, 0.14353448275862069em) $) -- ($ (M-6-2.south west) + (2.9600000000000004em, 0.047844827586206895em) $) -- ($ (M-6-2.south west) + (3.7em, 0.047844827586206895em) $) -- ($ (M-6-2.south west) + (3.7em, 0em) $) -- cycle;
\draw[opacity=0.6,fill=OddColor]($ (M-6-3.south west) + (0em, 0em) $) -- ($ (M-6-3.south west) + (0.0em, 0.478448275862069em) $) -- ($ (M-6-3.south west) + (0.7400000000000001em, 0.478448275862069em) $) -- ($ (M-6-3.south west) + (0.7400000000000001em, 0.6698275862068966em) $) -- ($ (M-6-3.south west) + (1.4800000000000002em, 0.6698275862068966em) $) -- ($ (M-6-3.south west) + (1.4800000000000002em, 0.09568965517241379em) $) -- ($ (M-6-3.south west) + (2.2200000000000006em, 0.09568965517241379em) $) -- ($ (M-6-3.south west) + (2.2200000000000006em, 0.09568965517241379em) $) -- ($ (M-6-3.south west) + (2.9600000000000004em, 0.09568965517241379em) $) -- ($ (M-6-3.south west) + (2.9600000000000004em, 0.047844827586206895em) $) -- ($ (M-6-3.south west) + (3.7em, 0.047844827586206895em) $) -- ($ (M-6-3.south west) + (3.7em, 0em) $) -- cycle;
\draw[opacity=0.6,fill=OddColor]($ (M-6-4.south west) + (0em, 0em) $) -- ($ (M-6-4.south west) + (0.0em, 0.5741379310344827em) $) -- ($ (M-6-4.south west) + (0.7400000000000001em, 0.5741379310344827em) $) -- ($ (M-6-4.south west) + (0.7400000000000001em, 0.5262931034482758em) $) -- ($ (M-6-4.south west) + (1.4800000000000002em, 0.5262931034482758em) $) -- ($ (M-6-4.south west) + (1.4800000000000002em, 0.14353448275862069em) $) -- ($ (M-6-4.south west) + (2.2200000000000006em, 0.14353448275862069em) $) -- ($ (M-6-4.south west) + (2.2200000000000006em, 0.09568965517241379em) $) -- ($ (M-6-4.south west) + (2.9600000000000004em, 0.09568965517241379em) $) -- ($ (M-6-4.south west) + (2.9600000000000004em, 0.047844827586206895em) $) -- ($ (M-6-4.south west) + (3.7em, 0.047844827586206895em) $) -- ($ (M-6-4.south west) + (3.7em, 0em) $) -- cycle;

\draw[opacity=0.6,fill=EvenColor]($ (M-7-2.south west) + (0em, 0em) $) -- ($ (M-7-2.south west) + (0.0em, 0.0em) $) -- ($ (M-7-2.south west) + (0.7400000000000001em, 0.0em) $) -- ($ (M-7-2.south west) + (0.7400000000000001em, 0.047844827586206895em) $) -- ($ (M-7-2.south west) + (1.4800000000000002em, 0.047844827586206895em) $) -- ($ (M-7-2.south west) + (1.4800000000000002em, 0.3349137931034483em) $) -- ($ (M-7-2.south west) + (2.2200000000000006em, 0.3349137931034483em) $) -- ($ (M-7-2.south west) + (2.2200000000000006em, 0.6219827586206896em) $) -- ($ (M-7-2.south west) + (2.9600000000000004em, 0.6219827586206896em) $) -- ($ (M-7-2.south west) + (2.9600000000000004em, 0.38275862068965516em) $) -- ($ (M-7-2.south west) + (3.7em, 0.38275862068965516em) $) -- ($ (M-7-2.south west) + (3.7em, 0em) $) -- cycle;
\draw[opacity=0.6,fill=EvenColor]($ (M-7-3.south west) + (0em, 0em) $) -- ($ (M-7-3.south west) + (0.0em, 0.0em) $) -- ($ (M-7-3.south west) + (0.7400000000000001em, 0.0em) $) -- ($ (M-7-3.south west) + (0.7400000000000001em, 0.0em) $) -- ($ (M-7-3.south west) + (1.4800000000000002em, 0.0em) $) -- ($ (M-7-3.south west) + (1.4800000000000002em, 0.09568965517241379em) $) -- ($ (M-7-3.south west) + (2.2200000000000006em, 0.09568965517241379em) $) -- ($ (M-7-3.south west) + (2.2200000000000006em, 0.6219827586206896em) $) -- ($ (M-7-3.south west) + (2.9600000000000004em, 0.6219827586206896em) $) -- ($ (M-7-3.south west) + (2.9600000000000004em, 0.6698275862068966em) $) -- ($ (M-7-3.south west) + (3.7em, 0.6698275862068966em) $) -- ($ (M-7-3.south west) + (3.7em, 0em) $) -- cycle;
\draw[opacity=0.6,fill=EvenColor]($ (M-7-4.south west) + (0em, 0em) $) -- ($ (M-7-4.south west) + (0.0em, 0.0em) $) -- ($ (M-7-4.south west) + (0.7400000000000001em, 0.0em) $) -- ($ (M-7-4.south west) + (0.7400000000000001em, 0.047844827586206895em) $) -- ($ (M-7-4.south west) + (1.4800000000000002em, 0.047844827586206895em) $) -- ($ (M-7-4.south west) + (1.4800000000000002em, 0.3349137931034483em) $) -- ($ (M-7-4.south west) + (2.2200000000000006em, 0.3349137931034483em) $) -- ($ (M-7-4.south west) + (2.2200000000000006em, 0.478448275862069em) $) -- ($ (M-7-4.south west) + (2.9600000000000004em, 0.478448275862069em) $) -- ($ (M-7-4.south west) + (2.9600000000000004em, 0.5262931034482758em) $) -- ($ (M-7-4.south west) + (3.7em, 0.5262931034482758em) $) -- ($ (M-7-4.south west) + (3.7em, 0em) $) -- cycle;

\draw[opacity=0.6,fill=OddColor]($ (M-8-2.south west) + (0em, 0em) $) -- ($ (M-8-2.south west) + (0.0em, 0.047844827586206895em) $) -- ($ (M-8-2.south west) + (0.7400000000000001em, 0.047844827586206895em) $) -- ($ (M-8-2.south west) + (0.7400000000000001em, 0.047844827586206895em) $) -- ($ (M-8-2.south west) + (1.4800000000000002em, 0.047844827586206895em) $) -- ($ (M-8-2.south west) + (1.4800000000000002em, 0.28706896551724137em) $) -- ($ (M-8-2.south west) + (2.2200000000000006em, 0.28706896551724137em) $) -- ($ (M-8-2.south west) + (2.2200000000000006em, 0.478448275862069em) $) -- ($ (M-8-2.south west) + (2.9600000000000004em, 0.478448275862069em) $) -- ($ (M-8-2.south west) + (2.9600000000000004em, 0.5262931034482758em) $) -- ($ (M-8-2.south west) + (3.7em, 0.5262931034482758em) $) -- ($ (M-8-2.south west) + (3.7em, 0em) $) -- cycle;
\draw[opacity=0.6,fill=OddColor]($ (M-8-3.south west) + (0em, 0em) $) -- ($ (M-8-3.south west) + (0.0em, 0.0em) $) -- ($ (M-8-3.south west) + (0.7400000000000001em, 0.0em) $) -- ($ (M-8-3.south west) + (0.7400000000000001em, 0.0em) $) -- ($ (M-8-3.south west) + (1.4800000000000002em, 0.0em) $) -- ($ (M-8-3.south west) + (1.4800000000000002em, 0.2392241379310345em) $) -- ($ (M-8-3.south west) + (2.2200000000000006em, 0.2392241379310345em) $) -- ($ (M-8-3.south west) + (2.2200000000000006em, 0.478448275862069em) $) -- ($ (M-8-3.south west) + (2.9600000000000004em, 0.478448275862069em) $) -- ($ (M-8-3.south west) + (2.9600000000000004em, 0.6698275862068966em) $) -- ($ (M-8-3.south west) + (3.7em, 0.6698275862068966em) $) -- ($ (M-8-3.south west) + (3.7em, 0em) $) -- cycle;
\draw[opacity=0.6,fill=OddColor]($ (M-8-4.south west) + (0em, 0em) $) -- ($ (M-8-4.south west) + (0.0em, 0.047844827586206895em) $) -- ($ (M-8-4.south west) + (0.7400000000000001em, 0.047844827586206895em) $) -- ($ (M-8-4.south west) + (0.7400000000000001em, 0.047844827586206895em) $) -- ($ (M-8-4.south west) + (1.4800000000000002em, 0.047844827586206895em) $) -- ($ (M-8-4.south west) + (1.4800000000000002em, 0.2392241379310345em) $) -- ($ (M-8-4.south west) + (2.2200000000000006em, 0.2392241379310345em) $) -- ($ (M-8-4.south west) + (2.2200000000000006em, 0.478448275862069em) $) -- ($ (M-8-4.south west) + (2.9600000000000004em, 0.478448275862069em) $) -- ($ (M-8-4.south west) + (2.9600000000000004em, 0.5741379310344827em) $) -- ($ (M-8-4.south west) + (3.7em, 0.5741379310344827em) $) -- ($ (M-8-4.south west) + (3.7em, 0em) $) -- cycle;
\end{tikzpicture} 
\end{center}
\begin{flushleft}
\footnotesize{Questions were asked on a 5-point Likert scale where the minimum value of 0.0 corresponds to strongly disagree whilst the maximum value of 4.0 corresponds to strongly agree. Coloured bars represents the distribution of participant responses.}
\end{flushleft}
\end{figure}

In our prior work \cite{Kal2}, participants’ most disliked aspect of the assistant was the uncertainty in inferring the intent of the assistant and what the assistant expected of them during the study. The inclusion of feedback cues addresses the prior work’s limitations by alleviating the uncertainty in what the assistant is trying to achieve and what the assistant wants the pilot to do by signalling when a task is ready to be completed by displaying the green light. This is supported by a statistically significant difference being observed between the Assisted \& Assisted + Lights condition for questions 1, 2, 3 \& 4 in Table~\ref{ConditionPerceptionTable} under a 95\% confidence interval using a Welch’s t-test. 

\begin{table}[b]
\begin{center}
\captionof{table}{Final survey worded questions}
\label{WordedResponse}
\begin{tabular}{|m{10cm}|}
  \hline
  \textbf{1:} What part of the task did you have the most difficulties with?\\
  \hline
  \textbf{2:} What did you most like about each of the conditions?\\
  \hline
  \textbf{3:} What did you most dislike about each of the conditions?\\
  \hline
  \textbf{4:} Rank the conditions from most preferred (1) to least preferred (3)\\
  \hline
  \textbf{5:} What features or changes would you make to any of the conditions?\\
  \hline
  \textbf{6:} Describe any differences in flying strategy for each of the conditions\\
  \hline
  \textbf{7:} Any additional comments/observations you would like to add?\\
  \hline
\end{tabular}
\end{center}
\end{table}

\subsubsection{Participant Free-Form Response}
The complete list of free-form questions can be found in Table.~\ref{WordedResponse}. When asked about the most difficult aspect of the task, 53\% of participants who responded mentioned difficulties with tracking the orientation of the UAV and correctly mapping joystick commands to move the UAV in the intended direction. It was observed that participants often got confused with the orientation of the UAV causing them to fly in an unintended direction where common solutions to this problem were to “explore by picking a direction and see if it goes where expected” or to “rotate the drone to face the same way I was”. Compared to our prior work \cite{Kal2} where the yaw remained locked, the previous most difficult aspect of the user study as stated by 68\% of participants was difficulties in perceiving depth of the UAV, while in the current study only 14\% of participants mentioned difficulties involving depth. Participants flying in an unintended direction due to incorrect orientation estimation often led to flying out of bounds and triggering the safety system. A total of 119 safety system triggers were observed during the 3 assisted conditions compared to a single safety system trigger in our prior study \cite{Kal2}. 

When asked about their most preferred aspect for each of the three conditions, for the Assisted condition 37\% of participants most liked how the assistant automatically aligned the UAV to the platform as they “didn't have to worry about perfectly piloting the drone” and instead having to only approximately reach the intended platform. For the Assisted + Lights condition 62\% of participants most liked the confirmation and certainty the assistant provided by flashing the green light when a task was ready to be completed where participants “found this condition to be both intuitive and learnable” with the feedback it provided. For the Assisted + HoloLens condition 46\% of participants most liked the display of the UAV’s orientation as the “augmented reality addition makes the orientation of the robot easiest to track, therefore making the control easiest”, which addresses the most difficult aspect of the user study from question 1 in Table~\ref{WordedResponse}. A further 36\% of participants mentioned they liked how the floor projection made it easy to tell if the UAV was above a platform by observing the change in height of the projected vertical line.

When participants were asked about their most disliked aspect of the task for each of the three conditions, for the Assisted condition the most disliked aspect generally pertained to the lack of certainty in the UAV’s position, orientation, inspection image quality or what the assistant wanted the participant to do. For the Assisted + Lights condition the majority of participants gave no response. For the Assisted + HoloLens condition 29\% of participants left comments stating that the HoloLens was “uncomfortable”, “heavy” or “gives you a headache” when wearing it. A further 36\% of participants stated that using the HoloLens degraded their visual abilities due to “decreased contrast” from the darkened glass visor, the augmented reality overlay obscuring real-life features and an obstruction of their peripheral vision from the visor.

Information about participant free-form response analysis procedures can be found in Appendix~\ref{QualitativeProcedureAppendix}

\subsubsection{Condition Ranking}
Participants were asked to rank the three conditions from their most preferred (1) to least preferred (3), where the most preferred condition was the Assisted + Lights condition with an average ranking of 1.59, followed by the HoloLens condition with an average ranking of 1.96 then the Assisted condition at 2.44. Participants most preferred the Assisted + Lights condition due to the intuitive design and easy to interpret green light feedback which alleviated the uncertainties in task success. The Assisted + HoloLens condition was the second most preferred condition due to providing additional information that indicates the orientation of the UAV to aid the pilot in moving in their intended direction. However the HoloLens was reported as uncomfortable to wear and hindered the pilot’s natural vision.

\section{Outdoor Demonstration}\label{section_outdoor}
To demonstrate that the proposed approach is feasible in non-laboratory conditions without access to high-quality pose feedback from the Vicon motion capture system, an outdoor flight experiment was held. An Intel RealSense Tracking Camera T265 was fitted to the UAV for onboard pose estimation which the flight controller used for flight stabilisation. The UAV was not fitted with a GPS to emulate a GPS-denied environment. The software architecture remained identical as outlined in Section \ref{UserStudySection}, where all network inference continued to be computed using only the onboard companion computer.

\begin{figure}[t]
\centering
\includegraphics[width=1.0\columnwidth]{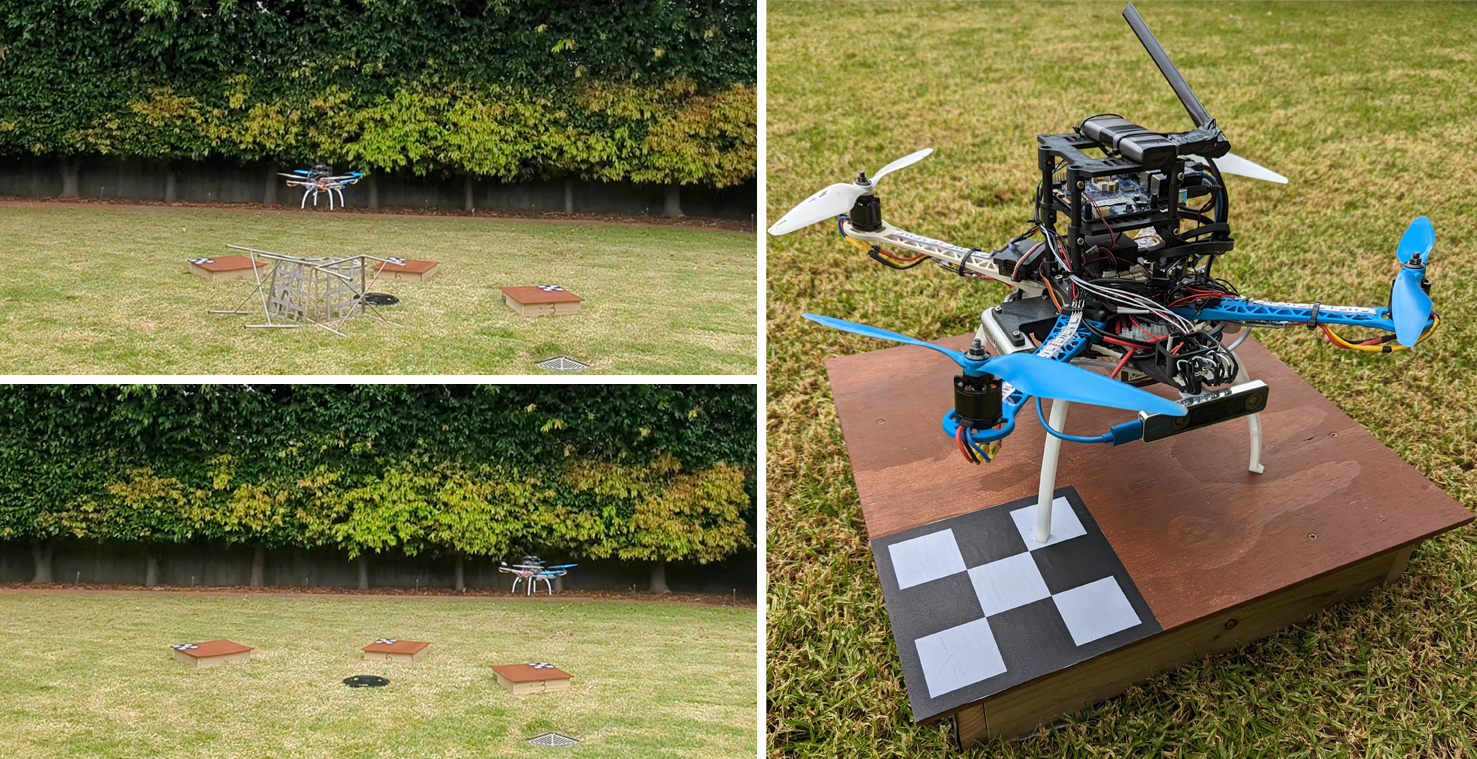}
\caption{(Top left) Pilot attempting to intentionally land in an unsafe location. (Bottom left) Pilot performing an inspection during the second experiment. (Right) UAV attached with the Intel RealSense Tracking camera T265 for onboard pose estimation.}
\label{OutdoorArena}
\end{figure}

Two outdoor experiments were held where the first experiment tested the assistant’s robustness in transferring to unfamiliar environments. The first author piloted the UAV in the assisted condition where they initially performed an inspection task followed by an intentional unsafe landing on top of an obstacle to observe the assistant’s behaviour. The second experiment aimed to measure the difference in performance to that of the indoor user study. For the second experiment the first author stood 10m behind the center platform and followed a standardised mission plan involving two inspection tasks followed by a final landing task. The mission was repeated a total of 20 times in both the assisted and unassisted condition. The outdoor experiment flight arena can be observed in Fig.~\ref{OutdoorArena}.

For the first experiment, the assistant was able to successfully abort the landing when the pilot intentionally attempted to land at an unsafe location by providing an opposing action to prevent the UAV from descending. For the second experiment, the average success rate for the landing and inspection task was [80\% \& 85\%] respectively compared to the unassisted condition which scored a success rate of [45\% \& 62.5\%]. Compared to the average assisted indoor user study results of [95.59\% \& 96.49\%], the lower success rate in the outdoor demonstration can be attributable to the poorer quality of pose estimation. During the experiment the UAV was observed to drift and be less responsive to pilot inputs, which degraded the performance of the assistant. 

\section{Discussion}
\subsection{Participant Behaviour to Task and Study Conditions}
Task performance across all three assisted conditions were similar to each other, sharing equally high success rates for the landing and inspection tasks due to the assistant being identical in each condition.
Compared to our prior work \cite{Kal2}, a greater diversity of flight strategies were observed in how participants approached the goal platform, primarily in how they accounted for the rotation of the UAV. Four fundamental strategies were observed to account for yaw when approaching a goal platform: (i) initially align the UAV with the ego-centric view so that the relative coordinate frame of the UAV is identical to that of the participant; (ii) initially align the UAV to face the target platform such that the participant only has to push the joystick forward to approach the platform; (iii) the participant would command the UAV to fly forward whilst adjusting the yaw to move to the target platform similar to driving a car and (iv) the participant would make no yaw adjustments and attempt to provide relative target velocity inputs to approach the platform. Although not all approach strategies were modelled in the simulated user model, it did not degrade the performance of the assistant in achieving task success.

Due to the diverse approach strategies, each mission was separated into approach, landing and inspection tasks in order to assess the impact each of the assisted conditions has on the landing and inspection tasks, without being influenced by the participant’s chosen approach strategy.  
Participants in the Assisted + Lights condition completed the inspection task more rapidly by requesting an image to be taken earlier due to the green light feedback from the assistant. Participants were not explicitly told the meaning behind the light feedback provided by the assistant, but were quickly able to learn that when the assistant displays the green light they are in a good position to complete the task by either requesting an image or descending. However it was noted that a specific mission with a short initial approach distance and the UAV already being correctly aligned to the goal platform would cause the assistant to require more time to display the green light as there was insufficient information about the intent of the participant. This led to some participants hovering above the platform, waiting for the assistant's signal before descending where they would have had otherwise attempted to land in an alternative condition.

Despite the Assisted + HoloLens being ranked as the second most preferred condition, participants’ free-form responses to the condition showed a varied perception, where some stated they enjoyed it whilst others expressed a strong dissatisfaction. The main sources of dissatisfaction were the weight and physical discomfort experienced, especially from participants who required prescription glasses. Other concerns from participants were the loss of calibration accuracy between the HoloLens and Vicon motion capture system, where some participants stated that the vertical floor projection line lost its centered alignment to the UAV and would display at the legs of the UAV. This issue would be further exacerbated in conditions without accurate motion capture feedback and limits the potential deployment of the HoloLens in practical UAV applications. 

Practical deployment of the HoloLens requires relative localisation between the UAV and HoloLens which is implemented in \cite{UAV_AR2}. An additional concern of practical implementation of the HoloLens in real-use scenarios is participants’ change in flight strategy. Some participants solely focused on the horizontal line projected onto the floor whilst flying, pretending they were driving a remote-controlled car. This proposed flight strategy may be optimal for controlled environments however could be dangerous in uncontrolled environments as it ignores potential overhanging and airborne obstacles.

\subsection{Learning and Training Efficiencies}
The success of the proposed approach can be attributable to three main factors: (i) a focus on training efficiency, (ii) improvements made to the base reinforcement algorithm TD3 and (iii) simulated user modeling in reacting to the assistant. Two prominent sources of training efficiency improvements implemented were the reduction of GPU workload when generating input images for policy exploration and the reduction in policy optimisation computation time by reusing LSTM cell embeddings. GPU workload reduction led to an increased exploration output by a factor of 3 compared to our prior work \cite{Kal2}, where increasing exploration further resulted in the training procedure becoming CPU bound. Sharing computed LSTM cell embeddings during policy optimisation between the actor and critic led to requiring 70\% of the computation time to perform a single policy training iteration. The aforementioned efficiency improvements led to an approximate total training time of 3 days per model which allowed for frequent reward structure modification to optimise the performance and piloting experience when transferring the model to the physical UAV.

Further improvements to training efficiency can be attained by pre-loading the replay buffer with user generated examples of good assistant behavior. Such examples minimize the exploration needs of the actor by providing a suitable foundation policy to improve from. This approach was found to reduce the time required to learn a policy during the testing and development phase where a small dataset of human generated assistant actions were collected from assisting the simulated user. This dataset was only used during development and was omitted for all models in the reported results.

The second factor for model success were the improvements made to the reinforcement learning algorithm TD3 by including a Temporal-Encoder network to share an LSTM embedding between the actor, critic and the introduced Temporal-Decoder. The Temporal-Decoder used supervised learning to forcefully embed information that can only be attained through successive temporal observations. Without the Temporal-Decoder it was empirically observed that the actor would often take a picture whilst the simulated user’s image request action was directly observable within the input state vector, regardless of image quality due to the actor not being able to remember whether an image was requested or not. This is further observed where models without the Temporal-Encoder would demonstrate suboptimal or inconsistent behaviour in displaying the red/green lights by either sporadically flashing them, continue to display them after a task is complete or chose to ignore light feedback completely. Sharing the Temporal-Encoder also resulted in fewer training iterations to reach convergence. However the increased performance with temporal dependant tasks may be due to increased amount of optimisation iterations being performed on the Temporal-Encoder. For the standard implementation of TD3, the Temporal-Encoder receives 1 actor optimisation per 2 training iterations whilst for the improved version a total of 5 Temporal-Encoder optimisations were performed every 2 training iterations.

Compared to our prior work \cite{Kal2} which successfully implemented the baseline TD3 reinforcement learning algorithm, the current study demonstrated that the standard TD3 model gave unsatisfactory results when task complexity was increased. In \cite{Kal2} the model was able to achieve a landing success rate of 80\% within the first 0.75 million training iterations and a final landing success rate of greater than 90\% after 2 million training iterations. However when transferring the TD3 algorithm to multi-task missions the model was only able to achieve a final landing success rate of 41.7\% despite being trained for twice as long for 4 million training iterations. This poorer performance can be attributable to the increased complexity in state-transitions with respect to the actor’s output due to the introduction of yaw control. 

Initial development focused on replicating the prior work’s \cite{Kal2} task of only landing the yaw locked UAV. The introduction of yaw control led to an increased observation of network instability with actor/critic loss spikes and catastrophic forgetting. This suboptimal behaviour is hypothesised to be due to sparser rewards from additional success requirements, increased dimensionality in actor’s output which has shown to cause instability in policy-gradient approaches for UAV landing tasks \cite{MovingPlatformDDPG, MovingPlatformDDPG2} and the increased difficulty in critic evaluation of the actor’s action for a given state due to the non-linearities introduced with rotation. However with the novel contributions to the TD3 reinforcement learning algorithm, the model was able to overcome the aforementioned difficulties and successfully complete multi-task missions.

\subsection{Simulated User Model}
The third factor of model success was the simulated user model and its ability to react to the assistant’s actions and communication feedback. Developing simulated user models for shared autonomy systems is challenging due to the difficulties associated with realistically replicating the broad range of policies pilots may implement. Implementing realistic reactions to the assistant’s actions is further difficult due to the unpredictability of the assistant throughout the training procedure. Parameterizing the simulated user model to include a parameter associated with the conformance to the assistant’s actions (\(\alpha\)) has shown success in our prior works \cite{Kal1, Kal2}. However teaching the assistant to provide meaningful communication feedback without an explicit reward structure whilst being easily interpretable by human pilots that are unfamiliar with the assistant is challenging. 
The simulated user model reacted to the red and green lights generated by the assistant by halting or progressing the current task, using commonly accepted social conventions behind the meaning of red and green lights.  Participants were able to quickly understand these signals and perceived the feedback as intuitive in the survey feedback. Having a hard requirement that samples of simulated users need the green state to progress the task forced the assistant to learn feedback cues to achieve task success. This hard requirement accurately reflects samples of human participants where the majority of pilots waited until the green light was displayed before progressing the task.

\subsection{Sim-to-Real Transfer}
The assistant was successful in transferring from simulation to reality as demonstrated in the user study described in section~\ref{UserStudySection}, despite being trained purely on simulated data. The assistant’s robustness to discrepancies between the simulated UAV’s dynamics and physical UAV’s dynamics are mitigated by not relying on low level control actions. Learning a policy tied to low level control actions such as individual motor thrusts increases the sensitivity to inaccuracies in the modeled dynamics and requires higher action output rates to compensate for the finer control required. As the proposed model outputs target velocities, it is the responsibility of the flight controller to bridge the gap in executing a trajectory that is compliant with the model’s output.

However given conditions that induce sub-optimal flight controller performance such as poor localization estimates, this gap becomes difficult to bridge. Poor flight controller performance was experienced during the outdoor demonstration as described in section~\ref{section_outdoor} due to the poor pose estimation methods employed in the GPS denied scenario. This caused the flight controller to execute delayed responses, leading to a discrepancy in the modeled dynamics and thus resulting in degraded model performance.

\subsection{Successful Assistance Strategy Design}
Across our UAV assistance studies \cite{Kal1}, \cite{Kal2} and the study in this manuscript, three common characteristics were prominently favored amongst participants: passiveness, clarity and intuitiveness. Passiveness is defined as the assistant providing as little assistance as necessary whilst not initiating important high-level decisions. Participants in \cite{Kal1} stated they disliked the assistant’s lack of passiveness when it would exert high levels of control to force a landing, despite the landing resulting in a success. This was rectified in the subsequent study \cite{Kal2} where participants’ most praised aspect of the assistant was that it was “non-invasive” and that they “did not notice it was there”. However for cases in which the drone was close to a platform, participants stated they disliked how the assistant would force the decision and perform an automated landing. 
In the current study, participants had a positive reception to the assistant’s automated yaw alignment. Unlike the dissatisfaction towards the forced automated landing, participants perceived this as passive behavior due to them maintaining control of the high-level objective whilst the low level control requiring precise yaw alignment was handled by the assistant. In summary, participants like to remain in control of the system regarding high level decisions such as deciding when to land, but prefer the assistant to correct for low level decisions involving precise perception and control as long as the user remains in charge of when and what action is being performed.

Future works should aim to maximize passiveness by focusing the assistance delivered to solve low level tasks that challenge users' perception and input control dexterity. Passiveness can be further improved by minimizing the difference in actions between the user and assistant. Further care is needed for reinforcement learning approaches, particularly in situations where a small amount of state transitions can lead to a large reward terminal state. Such situations can entice the agent to forego it’s passive behavior in favor of a large reward as seen in \cite{Kal2}.

Clarity is defined as the ease of understanding and perceiving the system and task. A common theme across the prior works \cite{Kal1, Kal2} was participants’ uncertainty in perceiving the intention of the assistant. Participants’ in \cite{Kal1} stated that they disliked the inconsistency in the behavior of the assistant, resulting in a lack of clarity of what to expect from the assistant. Participants’ in \cite{Kal2} disliked the inability to perceive the intent of the assistant claiming that “it was completely invisible”, leading to a sense of uncertainty in what was expected of them. In the current work these issues were solved by allowing the assistant to provide red / green light feedback to communicate its intent and the state of the system without negatively impacting passivity by taking over control of the system to execute intent displaying maneuvers.

Future works can promote clarity by ensuring that the system is predictable and consistent in its behavior. Providing feedback indicators that alert users of the current state or intent of the system can help reduce operator anxiety regarding the uncertainty of not knowing what to do or expect.

Intuitiveness is defined as the ease for users to interact, learn and understand how to complete tasks using the system. Across the current work and prior work \cite{Kal2} the most commonly acclaimed aspect of the assistant was its intuitiveness to use, requiring no training or guidance to work alongside. Participants praised the effortlessness of using the assistant due to simplifying the task to require only approximate high-level controls while fine precision control inputs were handled by the assistant. However the addition of yaw control was found to be unintuitive for novice pilots due to the unfamiliarity of operating under non-egocentric conditions. Despite the intentional omission of details surrounding the red / green light feedback condition, participants were able to quickly learn the meaning behind the assistant’s feedback due to following standardized conventions.

Future works can aim to maximize intuitiveness by alleviating the need for users to manage low level precise controls over the system, instead focusing on delivering general high-level commands. Extra care needs to be taken to simplify control schemes and avoid out-of-body unaligned coordinate frames due to being unintuitive to use without training. 
Future works should also aim on implementing well established conventions into their system so that users can draw upon prior knowledge to ease the learning process.

\section{Conclusion}
In this work we propose a shared autonomy approach that assists pilots in successfully completing multi-task missions comprised of inspection and landing tasks. We implement an AI assistant trained using our improved implementation of the deep reinforcement learning algorithm TD3 to augment the pilot’s actions, alongside supplementary information from red/green light feedback cues and augmented reality using the HoloLens to assist pilots. The assistant is comprised of three modules: a perception module responsible for encoding visual information onto a compressed latent vector, a policy module responsible for providing control outputs that are averaged with the pilot’s actions to control the UAV and an information augmentation module responsible for providing information about the state of the UAV and provide feedback to the pilot. The assistant was trained concurrently in simulation against a population of simulated users defined using a parametric model. The assistant was able to further influence the behaviour of simulated users by providing red/light feedback cues to entice certain actions.

A physical user study ($n=29$) was held to validate the proposed assistant and to assess participants’ perception to the proposed supplementary information schemes. Participants were tasked with completing 6 multi-task missions for each condition by taking multiple pictures of the nine specified platforms after which were tasked with safely landing on a specified platform. The assistant raised task success rate from [16.67\% \& 56.06\%] respectively in the unassisted condition to [95.59\% \& 96.49\%] in the assisted condition. The red/green light feedback cues were found to reduce the time required by participants to complete the inspection task by 19.41\% and the distance traveled by 17.80\%, whilst being their most preferred condition due to being perceived as intuitive to understand and provided the needed reassurance and confidence to the pilot about the current successfulness of the task. 

An outdoor experiment was held where the UAV flew in a GPS-denied environment. The assistant attained a success rate for the landing and inspection task of [80\% \& 85\%] respectively compared to [45\% \& 62.5\%] in the unassisted condition and was able to successfully transfer to unseen conditions without the need for motion capture feedback.


\bibliographystyle{ACM-Reference-Format}
\bibliography{sample-base}

\appendix

\section{Perception Module Implementation} \label{CM-SVAE_Implementation}
Implementation of the CM-SVAE architecture follows our previous implementation \cite{Kal2} of the CM-VAE architecture where two downwards facing RGB-D cameras situated to the left \(x_{L}\) = [\(x_{LRGB}\), \(x_{LD}\)] and right \(x_R\) = [\(x_{RRGB}\), \(x_{RD}\)] of the drone are used. Both \(x_L\) and \(x_R\) alongside their simplified versions \(\bar{x}_L\) and \(\bar{x}_R\) are passed through a Siamese feature extraction layer \(F_{DN}\) using DroNet \cite{DroNet}. The resultant vectors are concatenated and encoded onto the latent embedding of width 24 with encoder \(q_{RGBD}\) i.e.  \([\mu, \sigma ^2 ] = q_{RGBD}( [F_{DN}(x_{L}), F_{DN}(x_R)] )\) and \([\bar{\mu}, \bar{\sigma} ^2 ] = q_{RGBD}( [F_{DN}(\bar{x}_{L}), F_{DN}(\bar{x}_R)] )\). The CM-SVAE contains three output modalities: the simplified grayscale image \(\bar{y}_G\), the simplified depth map \(\bar{y}_D\) and segmentation map \(\bar{y}_S\) which denotes areas which are safe and unsafe to land. To reconstruct \(\bar{y}_G\), \(\bar{y}_D\) and \(\bar{y}_S\), a sample is taken from the latent embedding \(z \sim \mathcal{N}(\mu, \sigma^2)\) \& \(\bar{z} \sim \mathcal{N}(\bar{\mu}, \bar{\sigma}^2)\), which is subsequently fed through decoders \(p_G\), \(p_D\) and \(p_S\): \(\hat{\bar{y}}_G = p_G([z,\bar{z}])\), \(\hat{\bar{y}}_D = p_D([z,\bar{z}])\) and \(\hat{\bar{y}}_S = p_S([z, \bar{z}])\). 

In order to create a consolidated latent representation instead of two separate latent representations for \(x_L\) and \(x_R\), we implement a camera projection model that fuses the field of view of both cameras when reconstructing output images \(\bar{y}_G\), \(\bar{y}_D\) and \(\bar{y}_S\). The camera projection model in \cite{Kal2} is modified to account for differences in the position and orientation of the RGB-D cameras. We use input images of size 210w \(\times\) 120h with 20 intermediate cameras (\(N_c\) = 20), to create a combined output image size of 210w \(\times\) 140h which is subsequently down-sampled to 96w \(\times\) 64w.

\begin{figure}[b]
\centering
\includegraphics[width=1.0\columnwidth]{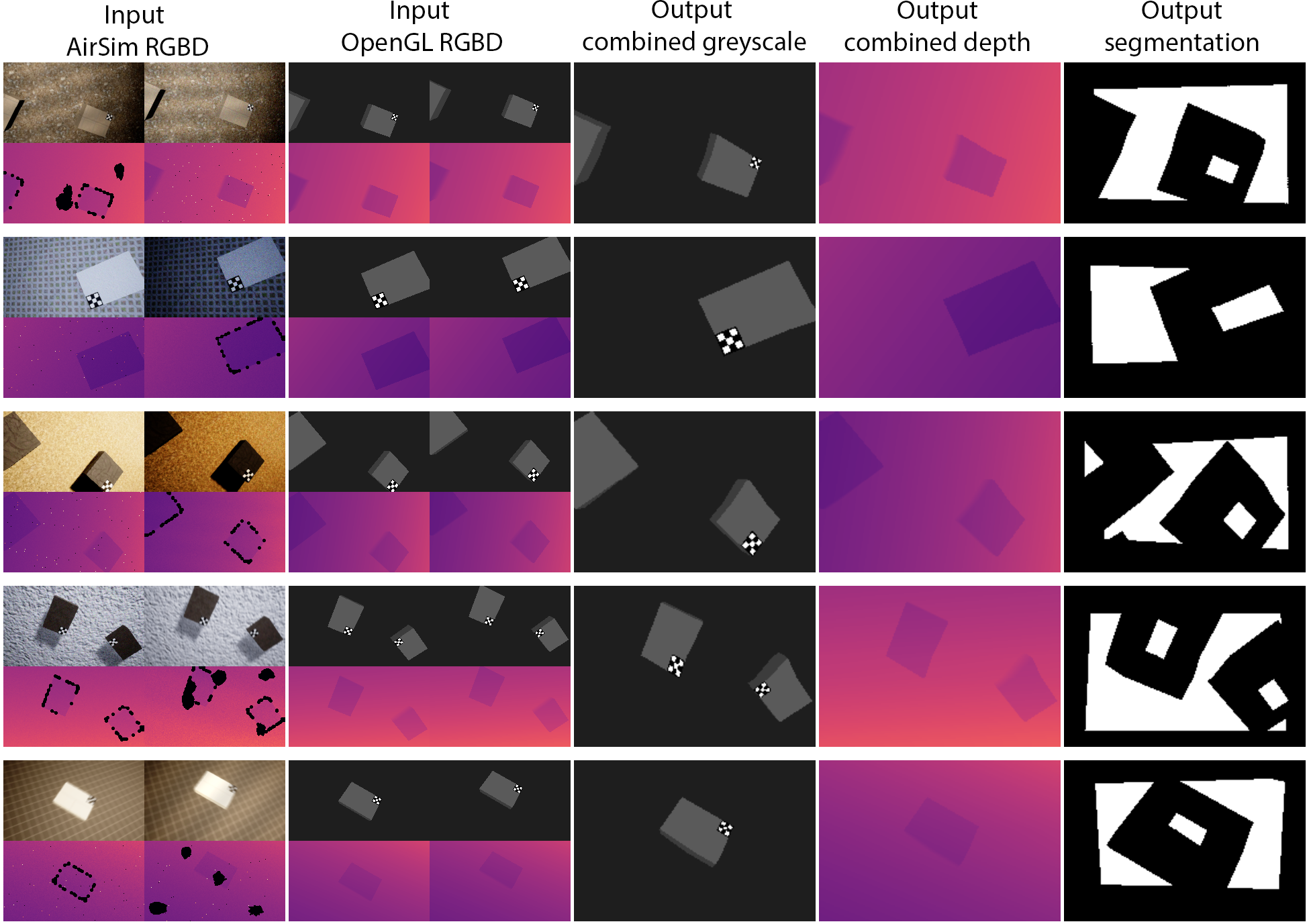}
\caption{Example images used to train the CM-SVAE architecture}
\label{CMSVAETrainingImages}
\end{figure}

High visual fidelity images \(x_L\) \& \(x_R\) are rendered with the AirSim \cite{AirSim} Unreal Engine plugin, while light-weight image rendering for \(\bar{x}_L\) \& \(\bar{x}_R\) uses OpenGL. To generate a scene, \(N_p\) platforms are randomly placed with random orientations such that adjacent platforms are not within a minimum distance threshold. For \(x_L\) \& \(x_R\), each generated scene has their lighting conditions altered and textures are randomly sampled from a database of 500 materials for the platforms, walls and floor. For \(\bar{x}_L\) \& \(\bar{x}_R\) lighting effects are disabled where the platforms, walls and floor are given a default colour which persists over all generated scenes. For each generated scene, a total of 10 left and right input RGB-D images are taken from both the Unreal Engine and OpenGL renderers in identical randomly generated poses. The output images \(\bar{y}_G\), \(\bar{y}_D\) and \(\bar{y}_S\) are generated using only the OpenGL renderer. As the images generated using AirSim are intended to be reflective of those from real RGB-D cameras, a noise generating function \(G_n\) is used to mimic sensor noise present in real-life cameras. For greater domain randomisation and robustness against sensor noise, \(G_n\) is applied to input batches of clean AirSim images which are dynamically noised during training by parallel worker threads. Example generated training images can be seen in Fig.~\ref{CMSVAETrainingImages}

\begin{figure}
\centering
\includegraphics[width=1.0\columnwidth]{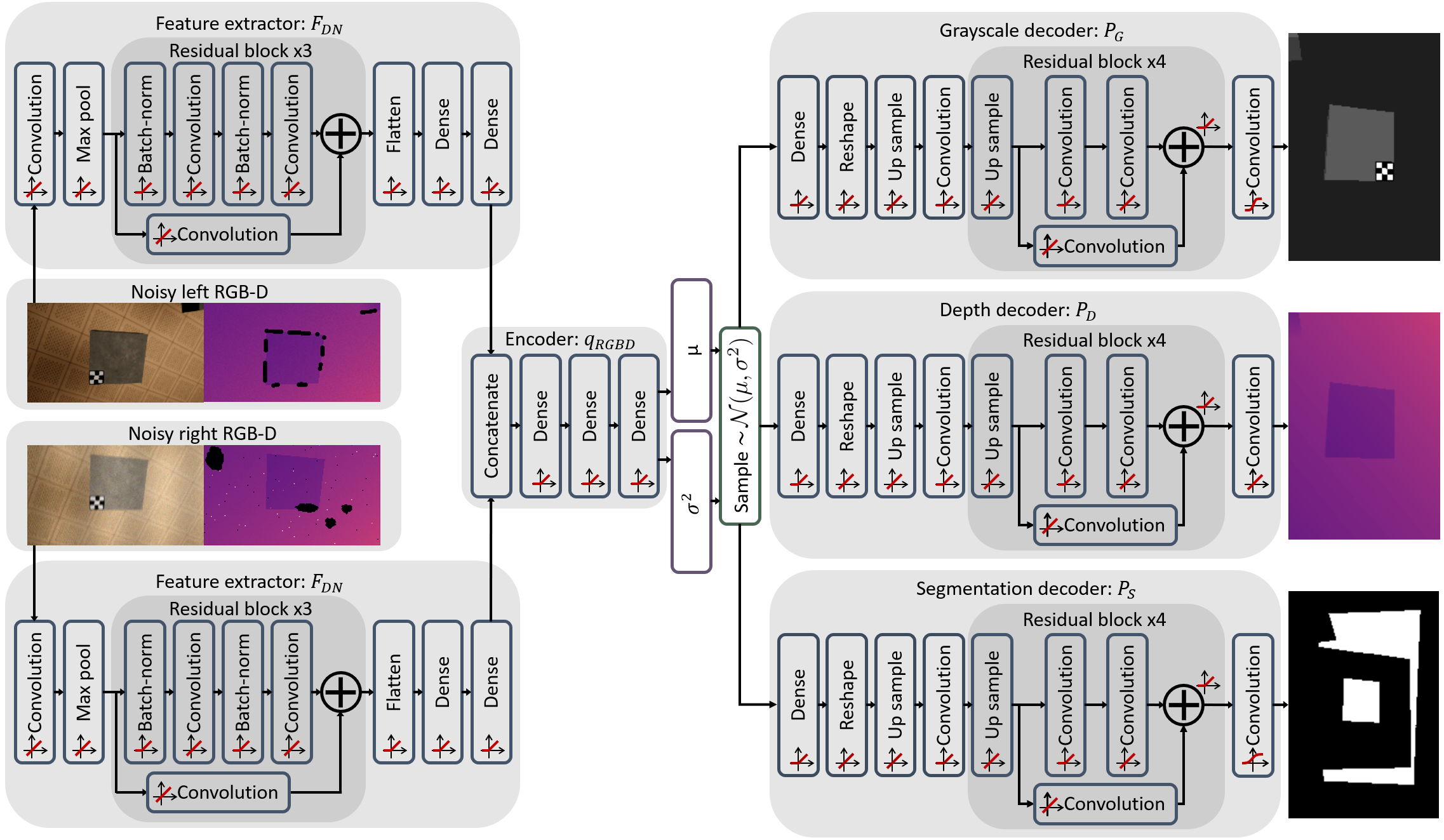}
\caption{Overview of the CM-SVAE architecture}
\label{CMSVAEModel}
\end{figure}

Losses used to train the perception module can be classified into two categories: (i) those that aim to embed information onto the latent vector and (ii) those that aim to condition the latent vector. For (i) we used the mean squared error between the true grayscale images and the estimated grayscale images: \(L_{G}=MSE(\bar{y}_G, \hat{\bar{y}}_G)\), the mean squared error between the true depth map and estimated depth map: \(L_{D}=MSE(\bar{y}_D, \hat{\bar{y}}_D)\) and the binary cross entropy loss between the true segmentation map and estimated segmentation map: \(L_{Seg}=BCE(\bar{y}_S, \hat{\bar{y}}_S)\). For (ii) we use the Kullback-Leibler divergence to maintain the desired latent distribution of \(\mu = 0\) and \(\sigma^2 = 1\) (\(L_{KL})\) and the cosine similarity loss to maintain an identical latent embedding between the AirSim and OpenGL rendered images \(L_{Sim}=\frac{\mu \cdot \bar{\mu}}{\norm{\mu} \norm{\bar{\mu}}}\). For each loss: \(L_{G}\), \(L_{D}\), \(L_{Seg}\), \(L_{KL}\) and \(L_{Sim}\) we give a weighting of 3.0, 1.0, 0.5, 4.0 and 1.0 respectively. An overview of the CM-SVAE architecture can be seen in Fig.~\ref{CMSVAEModel}.

\section{Policy Module Training} \label{PolicyImplementation}
\begin{figure}
\centering
\includegraphics[width=0.95\columnwidth]{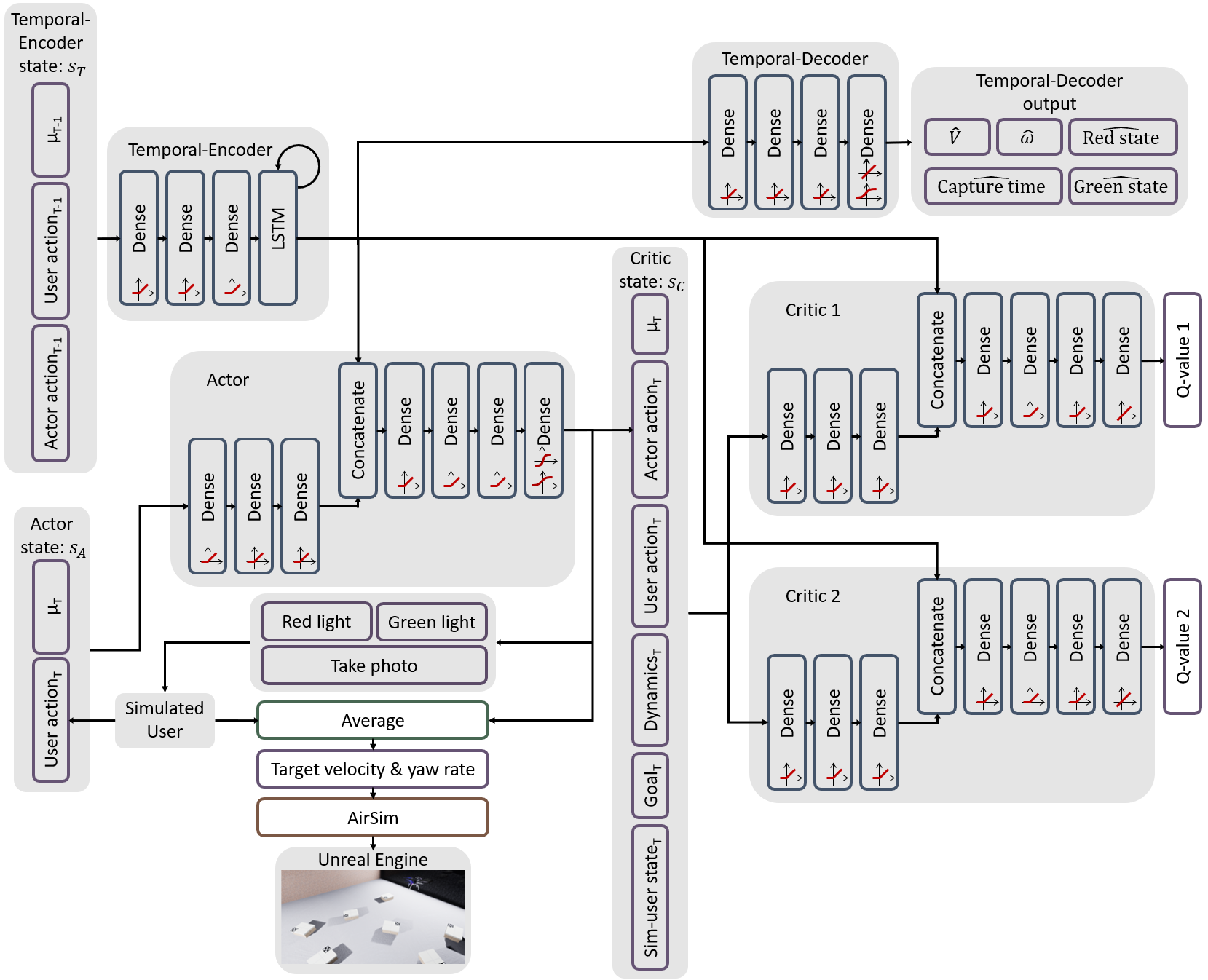}
\caption{Overview of assistant training architecture.}
\label{SharedTD3}
\end{figure}

An overview of the policy module training architecture can be seen in Fig.~\ref{SharedTD3}. For the input states \(s=[s_T, s_A, s_C]\) associated with the Temporal-Encoder, actor and critic respectively, \(s_T\) is comprised of three inputs: the previous mean latent vector from encoder \(q_{RGBD}\), the user’s previous action and the assistant’s previous action. The user’s action includes their target XYZ velocity, yaw rate and take-photo action if an image was requested in the given state. The assistant’s action includes its target XYZ velocity, yaw rate, take-photo action, red and green light display actions.
\(s_A\) is comprised of two components: the current mean latent vector from encoder \(q_{RGBD}\) and the user’s current action.
As \(s_C\) is required in training only, it can be supplemented with privileged information that can only be attained during simulation which was found to have the greatest impact in network performance in our prior work \cite{Kal2}. \(s_C\) comprises of six components: the UAV dynamics, user’s action, actor's action, current goal information, simulated user’s state and the mean latent vector from encoder \(q_{RGBD}\). The UAV dynamics includes the current position, yaw, linear and angular velocities of the UAV. The current goal information includes the relative position and yaw of the goal platform with respect to the current UAV’s pose, the type of mission the simulated user is attempting to complete (landing or inspection) and the physical dimensions of the goal platform. The simulated user’s state information includes their patience, green and red light state, as well as what the simulated user’s current objective is (approach platform / landing / inspecting).

The reward structure used to compute the reward for a given state transition can be categorised into three components: landing rewards, inspection rewards and cooperation rewards where the total reward for a given state transition is defined as: 
\begin{equation}\label{RewardEq}
\begin{aligned}
R = R_\mathrm{Landing} +R_\mathrm{Inspection} + R_\mathrm{Cooperation}.
\end{aligned}
\end{equation}  

\(R_\mathrm{Landing}\) contains four terms: landing success \(R_\mathrm{LS}\) and the landing velocities along the horizontal plane \(R_\mathrm{HVel}\), vertical direction \(R_\mathrm{VVel}\) and yaw rate \(R_\mathrm{YVel}\). \(R_\mathrm{LS}\) is a discrete reward which takes on a value of -1 given a successful landing or a value of 1 given a failed landing. A successful landing is defined as one where the UAV lands with all four legs on the correct goal platform, where the intent of the user is to land and not inspect the goal platform and that the difference between the yaw of the drone to that of the platform’s orientation is within a threshold  of $\pm 20$ degrees. For \(R_\mathrm{HVel}\), \(R_\mathrm{VVel}\) and \(R_\mathrm{YVel}\), the penalty is based on the velocity term squared scaled by the height \(H\) of the UAV above the platform up to a threshold height \(H_T\):
\begin{equation}
R_\mathrm{iVel} = 
    \begin{cases}
        \frac{(H_T - H)}{H_T}  v_i^2  &\text{if \(H < H_{T}\)} \\
        0                                  &\text{else}
    \end{cases},
\end{equation}
where \(v_i\) represents the specific velocity direction term. The full equation for landing reward term is: \(R_\mathrm{Landing} = k_0 R_\mathrm{LS} + k_ 1 R_\mathrm{HVel} + k_2 R_\mathrm{VVel} + k_3 R_\mathrm{YVel} \) where the weight of the scaling coefficients \(k_{0}\) \textendash \ \(k_{3}\) are given values of 12.0, 30.0, 5.0 \& 15.0 respectively.

\(R_\mathrm{Inspection}\) comprises of four terms: inspection pending time penalty \(R_\mathrm{IT}\),  inspection success \(R_\mathrm{IS}\), inspection velocity \(R_\mathrm{IVel}\) and inspection angular velocity \(R_\mathrm{IA}\). \(R_\mathrm{IT}\) is included to promote the assistant in taking images promptly after the pilot has requested one and is computed as: \( R_\mathrm{IT}=1.07^{T}\), where \(T\) is the number of timesteps that have elapsed since the pilot has requested an image to be taken. For \(R_\mathrm{IS}\), a success is considered if all four corners of the goal platform are present within the combined image representation and that the relative yaw error between the UAV and goal platform is within a threshold of $\pm 20$ degrees. \(R_\mathrm{IS}\) is calculated as: 
\begin{equation}
R_\mathrm{IS}=
    \begin{cases} 
        \frac{-P}{W  H} &\text{if success}\\ 
        1 &\text{else} 
    \end{cases},
\end{equation}
where \(P\) is the total pixels within the combined image representation that correspond to a platform while \(W\) and \(H\) are the width and height of the combined image. \(R_\mathrm{IS}\) is scaled by the proportion of platform pixels within an image to ensure the assistant takes a close-up image in order to capture as much detail as possible. \(R_\mathrm{IVel}\) and \(R_\mathrm{IA}\) are included to ensure the assistant takes a photo whilst the UAV is stationary to avoid motion blur and is calculated as \(R_\mathrm{IVel}=V^2\) and \(R_\mathrm{IA}=\omega^2\), where \(V\) and \(\omega\) are the corresponding linear and angular velocities. The full equation for the inspection reward term is: \(R_\mathrm{Inspection}=  k_4 R_\mathrm{IT} + k_5 R_\mathrm{IS} + k_6 R_\mathrm{IVel} + k_7 R_\mathrm{IA}\) where the weight of the scaling coefficients \(k_{4}\) \textendash \ \(k_{7}\) are given  values of 0.4, 50.0, 150.0 \& 150.0 respectively.

\(R_\mathrm{Cooperation}\) comprises of four terms: XYZ action difference between the pilot and assistant \(R_\mathrm{XYZ}\), the yaw rate action difference between the pilot and assistant \(R_\mathrm{Yaw}\), the red light action output of the assistant \(R_\mathrm{R}\) and the green light action output of the assistant \(R_\mathrm{G}\). \(R_\mathrm{R}\) and \(R_\mathrm{G}\) are given a value of 1 when the respective light is turned on, and 0 otherwise. The full equation for the cooperation reward term is: \(R_\mathrm{Cooperation}=  k_8 R_\mathrm{XYZ} + k_9 R_\mathrm{Yaw} + k_{10} R_\mathrm{R} + k_{11} R_\mathrm{G}\), where the scaling coefficients \(k_{8}\) \textendash \ \(k_{11}\) are given values of 0.45, 0.3, 0.07 \& 0.07 respectively.

For training the Temporal-Decoder we utilize five different losses to condition the Temporal-Encoder. 1) The mean squared error between the Temporal-Decoder’s estimated linear velocity and the UAV’s actual linear velocity: \(L_{LV} = MSE(\hat{V}, V)\). 2) The mean squared error between the Temporal-Decoder’s estimated angular velocity and the UAV’s actual angular velocity: \(L_{AV} = MSE(\hat{\omega}, \omega)\). 3) The mean squared error between the Temporal-Decoder’s estimate of the time since the pilot has requested an image yet an image has not been taken to that of the actual outstanding capture time: \(L_{OCT} = MSE(\hat{OCT}, OCT)\). 4) The binary cross entropy loss between the Temporal-Decoder’s estimate of whether the simulated user is currently in the red state and the simulated user’s red state variable: \(L_{RS} = BCE(\hat{RS}, RS)\). 5) The binary cross entropy loss between the Temporal-Decoder’s estimate of whether the simulated user is currently in the green state and the simulated user’s green state variable: \(L_{GS} = BCE(\hat{GS}, GS)\). The total loss for the Temporal-Decoder in training step (1) in Section \ref{Shared-TD3 Algorithm} is defined as: \(L_{TD} = k_{12}L_{LV} + k_{13}L_{AV} + k_{14}L_{OCT} + k_{15}L_{RS} + k_{16}L_{GS}\) where \(k_{12}-k_{16}\) are given values of 1.0, 1.0, 5.0, 1.5, 1.5 respectively.  

During training, the exploration process is parallelised with 24 concurrent simulated UAVs flying within a single Unreal Engine instance where the UAV simulation plugin AirSim has its simulation clock speed sped up by a factor of 2. An iteration is performed every 200ms in simulated time (100ms in real-life). For exploratory noise an Ornstein-Uhlenbeck process was used which was successively decayed after each epoch. 

The target XYZ velocity and yaw rate that the simulated UAV is to follow is calculated by averaging the simulated users and assistant’s actions, after which clipped normally distributed noise is added, with the variance of the normally distributed noise is derived from the current distance from the ground to emulate disturbances caused from the ground effect. After each epoch, policy module optimisation iterations are performed in accordance to the number of state transitions observed during the epoch using a batch size of 64. An epoch is considered complete when all 24 UAVs have completed their randomly generated mission list containing [1-3] tasks or if a total of 3-minutes of simulation time has elapsed. Training and model validation results are described in Section \ref{Result_TD3}.

\section{Learning effect regression result}\label{LearningEffectAppendix}
\begin{table}[!h]
\begin{center}
\captionof{table}{Learning effect regression model}
\newcommand\RegressionTableTwoFirstWidth{3.4}
\newcommand\RegressionTableTwoSecondWidth{1.2}
\setlength{\tabcolsep}{0.2em}
\label{RegressionSummaryLearning}
\definecolor{TableGray}{gray}{0.9}
\begin{tabular}{|p{\RegressionTableTwoFirstWidth cm}|c|c|c|c|}
  
  \hline
  \multicolumn{1}{|p{\RegressionTableTwoFirstWidth cm}|}{} & \multicolumn{1}{|p{\RegressionTableTwoSecondWidth cm}|}{\centering \(A\)} & \multicolumn{1}{|p{\RegressionTableTwoSecondWidth cm}|}{\centering \(AL\)} & \multicolumn{1}{|p{\RegressionTableTwoSecondWidth cm}|}{\centering \(AH\)} & \multicolumn{1}{|p{\RegressionTableTwoSecondWidth cm}|}{\centering \(CO\)}\\
  \hline
  \multicolumn{5}{|c|}{Target approach} \\ 
  \hline
  \multicolumn{1}{|p{\RegressionTableTwoFirstWidth cm}|}{\multirow{1}{*}{Time / init. dist.}}           & \multicolumn{1}{|c|}{\textbf{6.31}} & \multicolumn{1}{|c|}{\textbf{6.50}} & \multicolumn{1}{|c|}{\textbf{6.98}} & \multicolumn{1}{|c|}{\textbf{-0.82}}\\
  \rowcolor{TableGray}
  \multicolumn{1}{|p{\RegressionTableTwoFirstWidth cm}|}{\multirow{1}{*}{Dist. traveled / init. dist.}} & \multicolumn{1}{|c|}{\textbf{1.40}} & \multicolumn{1}{|c|}{\textbf{1.43}} & \multicolumn{1}{|c|}{\textbf{1.55}} & \multicolumn{1}{|c|}{\textbf{-0.11}} \\
  \hline
  \multicolumn{5}{|c|}{Landing task} \\ 
  \hline
  \multicolumn{1}{|p{\RegressionTableTwoFirstWidth cm}|}{\multirow{1}{*}{Time / init. dist.}}           & \multicolumn{1}{|c|}{\textbf{15.60}}  & \multicolumn{1}{|c|}{\textbf{16.17}}   & \multicolumn{1}{|c|}{\textbf{17.62}} & \multicolumn{1}{|c|}{\textbf{-1.83}} \\
  \rowcolor{TableGray}
  \multicolumn{1}{|p{\RegressionTableTwoFirstWidth cm}|}{\multirow{1}{*}{Dist. traveled / init. dist.}} & \multicolumn{1}{|c|}{\textbf{2.39}}  & \multicolumn{1}{|c|}{\textbf{2.54}}   & \multicolumn{1}{|c|}{\textbf{2.75}} & \multicolumn{1}{|c|}{\textbf{-0.21}} \\
  \hline
  \multicolumn{5}{|c|}{Inspection task} \\ 
  \hline
  \multicolumn{1}{|p{\RegressionTableTwoFirstWidth cm}|}{\multirow{1}{*}{Time / init. dist.}}           & \multicolumn{1}{|c|}{\textbf{12.31}}  & \multicolumn{1}{|c|}{\textbf{10.19}}   & \multicolumn{1}{|c|}{\textbf{13.43}} & \multicolumn{1}{|c|}{\textbf{-1.25}} \\
  \rowcolor{TableGray}
  \multicolumn{1}{|p{\RegressionTableTwoFirstWidth cm}|}{\multirow{1}{*}{Dist. traveled / init. dist.}} & \multicolumn{1}{|c|}{\textbf{1.77}}  & \multicolumn{1}{|c|}{\textbf{1.49}}   & \multicolumn{1}{|c|}{\textbf{2.11}} & \multicolumn{1}{|c|}{\textbf{-0.14}} \\
  \hline  
  
\end{tabular}
\end{center}
\begin{flushleft}
\footnotesize
Regression coefficients for the model \(Y = A + AL + AH + CO\). \\
Bolded values represent statistically significant values under a two-tailed t-test of significance \(\alpha = 0.01\).
\end{flushleft}
\end{table}

\section{Participant Free-Form Response Analysis Procedure}\label{QualitativeProcedureAppendix}
Participant qualitative responses were initially collated into individual documents for each of the questions asked in Table~\ref{WordedResponse}. An initial screening of responses was performed to discern shared themes amongst participants, where for each response a list of keywords was generated. During initial screening, the number of participants providing a valid response was recorded. Invalid responses that were excluded from the response count were those who left no comments or those who left comments which dismissed the question such as “N/A”, “None” or “No comment”. Initial keywords were tallied and concrete themes were generated using the most proliferate keywords.

Secondary screening was performed where each comment within a response was copied into an additional document reserved for each of the chosen themes discovered in the initial screening. Comments that refer to multiple themes were copied multiple times for each of the associated themes. Unassigned comments were assigned to a “miscellaneous” document. Responses in each keyword document were subsequently validated to ensure they share a common intent. Given a discrepancy in the shared intent of the responses or if common sub-themes were discovered, the chosen themes were modified to better reflect the responses and the secondary screening process was iteratively repeated until no changes were required.

The total unique participant responses for each theme was tallied and theme popularity was calculated as the percentage of participants commenting on the given theme over the total valid responses for the question.

\end{document}